\documentclass[preprint,12pt]{elsarticle}

\usepackage{comment}
\usepackage{amssymb}
\usepackage{amsmath}
\usepackage{url}
\usepackage{multirow}
\usepackage{lscape} 
\usepackage{longtable}
\usepackage{booktabs}
\usepackage{afterpage}
\usepackage{adjustbox}

\begin{document}

\begin{frontmatter}

\address[sycai]{Sycai Technologies SL, Scientific and Technical Department, Barcelona, Spain}
\address[upf]{BCN Medtech, Universitat Pompeu Fabra, Barcelona, Spain}
\address[uker1]{Universitätsklinikum Erlangen, Department of Radiology, Uniklinikum Erlangen, Erlangen, Germany}
\address[uker2]{University Hospital Erlangen, Imaging Science Institute, Erlangen, Germany}   
    
\author[sycai,upf]{Meritxell Riera-Marín}
\author[upf]{Sikha O K}
\author[sycai]{Júlia Rodríguez-Comas}
\author[uker1,uker2]{Matthias Stefan May}
         
\author[medig1,medig2]{Zhaohong Pan}
\author[medig1]{Xiang Zhou}
\author[medig1]{Xiaokun Liang}

\author[praecision]{Franciskus Xaverius Erick}
\author[praecision]{Andrea Prenner}

\author[breizhseg]{Cédric Hémon}
\author[breizhseg]{Valentin Boussot}
\author[breizhseg]{Jean-Louis Dillenseger}
\author[breizhseg]{Jean-Claude Nunes}

\author[abdul1]{Abdul Qayyum}
\author[abdul2]{Moona Mazher}
\author[abdul2]{Steven A Niederer}

\author[bcnaim1]{Kaisar Kushibar}
\author[bcnaim1]{Carlos Martín-Isla}
\author[bcnaim2]{Petia Radeva}
\author[bcnaim1,icrea]{Karim Lekadir}

\author[cai4cai]{Theodore Barfoot}
\author[cai4cai]{Luis C. Garcia Peraza Herrera}
\author[cai4cai]{Ben Glocker}
\author[cai4cai]{Tom Vercauteren}

\author[predicted1]{Lucas Gago}
\author[predicted2]{Justin Englemann}

\author[uker1]{Joy-Marie Kleiss}
\author[rad1,rad2]{Anton Aubanell}
\author[rad3]{Andreu Antolin}

\address[rad1]{Hospital de Sant Pau i la Santa Creu, Diagnostic Imaging Department, Barcelona, Spain}
\address[rad2]{Institut de Recerca Sant Pau - Centre CERCA, Advanced Medical Imaging, Artificial Intelligence, and Imaging-Guided Therapy Research Group, 
                Barcelona, Spain}
\address[rad3]{Hospital Universitari Vall d'Hebron, Department of Radiology, Institut de Diagnòstic per la Imatge (IDI), Barcelona, Spain}

\author[sycai]{Javier García-López}
\author[upf,icrea]{Miguel A. González Ballester}
\author[tecnalia]{Adrián Galdrán}

\address[tecnalia]{TECNALIA, Basque Research and Technology Alliance (BRTA), Bizcaia, Spain}

\address[medig1]{Shenzhen Institute of Advanced Technology, Chinese Academy of Sciences, Shenzhen, China}
\address[medig2]{University of Chinese Academy of Sciences, Beijing, China}

\address[praecision]{Friedrich-Alexander-Universität Erlangen-Nürnberg (FAU), Erlangen, Germany}  

\address[abdul1]{National Heart and Lung Institute, Faculty of Medicine, Imperial College London, London, United Kingdom}
\address[abdul2]{Centre for Medical Image Computing, Department of Computer Science, University College London, London, United Kingdom}

\address[bcnaim1]{Barcelona Artificial Intelligence in Medicine Lab (BCN-AIM), Facultat de Matemàtiques i Informàtica, Universitat de Barcelona,
             Barcelona, Spain}
\address[bcnaim2]{IBA, Facultat de Matemàtiques i Informàtica, and Institute of Neuroscience, Universitat de Barcelona, Barcelona, Spain}

\address[cai4cai]{King's College London (KCL), London, United Kingdom}

\address[predicted1]{Universitat de Barcelona (UB), Barcelona, Spain}
\address[predicted2]{University of Edinburgh, Edinburgh, United Kingdom}

\address[breizhseg]{Université de Rennes 1, CLCC Eugène Marquis, and INSERM UMR 1099 LTSI, Rennes, France}
                        
\address[icrea]{Institució Catalana de Recerca i Estudis Avançats (ICREA), Barcelona, Spain}

\title{Calibration and Uncertainty for multiRater Volume Assessment in multiorgan Segmentation (CURVAS) challenge results}

%% Abstract
\begin{abstract}
Deep learning (DL) has become the dominant approach for medical image segmentation, yet ensuring the reliability and clinical applicability of these models requires addressing key challenges such as annotation variability, calibration, and uncertainty estimation. This is why we created the Calibration and Uncertainty for multiRater Volume Assessment in multiorgan Segmentation (CURVAS), which highlights the critical role of multiple annotators in establishing a more comprehensive ground truth, emphasizing that segmentation is inherently subjective and that leveraging inter-annotator variability is essential for robust model evaluation. Seven teams participated in the challenge, submitting a variety of DL models evaluated using metrics such as Dice Similarity Coefficient (DSC), Expected Calibration Error (ECE), and Continuous Ranked Probability Score (CRPS). By incorporating consensus and dissensus ground truth, we assess how DL models handle uncertainty and whether their confidence estimates align with true segmentation performance. Our findings reinforce the importance of well-calibrated models, as better calibration is strongly correlated with the quality of the results. Furthermore, we demonstrate that segmentation models trained on diverse datasets and enriched with pre-trained knowledge exhibit greater robustness, particularly in cases deviating from standard anatomical structures. Notably, the best-performing models achieved high DSC and well-calibrated uncertainty estimates. This work underscores the need for multi-annotator ground truth, thorough calibration assessments, and uncertainty-aware evaluations to develop trustworthy and clinically reliable DL-based medical image segmentation models.
\end{abstract}

%% Keywords

\end{frontmatter}



\section{Introduction}\label{introduction}
Precise image segmentation is a central topic of current medical image analysis and its accurate assessment is crucial for early diagnosis, personalized treatment planning, and outcome prediction. 
Machine learning and artificial intelligence (AI) have advanced the analysis of complex imaging data, but reliable, public datasets and performance benchmarks remain essential for developing and validating deep learning (DL) models. 
A key challenge in this process is data uncertainty, particularly the impact of annotator disagreements on segmentation performance. 
Even experienced radiologists may disagree on anatomical boundaries, particularly in ambiguous regions, and precise delineation of such structures is critical.
Utilizing multiple annotations from different experts is a natural approach to addressing this challenge. 
By exploiting this information, models can be more robust and aware of data samples or image regions containing inherently ambiguous information.
However, some techniques rely on merging annotations with conventional methods such as label smoothing \cite{sudre_lets_2019,zhang_soft_2023,islam_spatially_2021}, or other methods like random label sampling \cite{jensen_improving_2019}, can be counterproductive, disregarding valuable information from annotator disagreements \cite{jungo_effect_2018}.
Moreover, while multi-rater annotations are sometimes incorporated during training, final evaluation typically relies on a single “gold standard,” which may fail to capture the variability inherent in multiple expert annotations, resulting in a less thorough and potentially skewed assessment.
This is problematic when datasets involve multiple annotators labeling different subsets of images, leading to biased evaluations. 
For example, models trained on one set of annotations may be assessed against a different set, reflecting subjective interpretations rather than predictive errors.
In clinical practice, such data can have a significant aspect in decision-making \cite{tuijn_reducing_2012}. 
No comprehensive benchmark of models leveraging multiple annotations in complex segmentation tasks has been conducted so far. 

% why we need uncertainty quantification - DONE
Modeling multi-rater variability falls within the scope of the area of uncertainty quantification and it plays a crucial role in medical image analysis.
Uncertainty reflects the degree of confidence in model predictions, arising from data variability, noise, and annotator disagreements. 
Accurate uncertainty estimates identify regions of low confidence, highlighting areas where clinicians should exercise caution. 
For example, Ng et al. \cite{ng2022estimating} employed uncertainty quantification for cardiac MRI segmentation quality control, using it for out-of-distribution detection or DeVries et al. \cite{devries2018leveraging} introduced a model integrating images, predicted segmentations, and uncertainty maps to estimate DSC scores and flag cases for expert review. 
Incorporating uncertainty into model evaluation enhances robustness and segmentation quality assessment \cite{abdar2021review} as well as aids clinical decision-making.
While much of the research in computational medical image analysis has traditionally focused on predictive performance, recent years have seen a growing interest in ensuring models provide meaningful uncertainty estimates \cite{lambert_trustworthy_2024}. 
A common measure of uncertainty \cite{bernhardt_failure_2022} is the predictive confidence, often calculated using the maximum softmax probability, used in the evaluation metrics of this challenge.
In recent years there has been an increasing focus in aligning this predictive confidence with accuracy, a concept known as calibration \cite{guo_calibration_2017, jensen_improving_2019}.
% why we need calibration - DONE
Calibration is essential for clinical trustworthiness, as model reliability must be assessed alongside segmentation accuracy to ensure a robust and confidence-aligned model.
However, calibration is a spectrum rather than a binary state.
A well-calibrated model provides confidence estimates that reflect true prediction reliability, but models may be underconfident, assigning low confidence to accurate predictions, or overconfident, with confidence exceeding actual accuracy. 
Evaluating miscalibration requires estimating accuracy at different confidence levels, which can be a complex task \cite{nixon_measuring_2019}. 
Understanding miscalibration patterns can guide model improvement and ensure safe clinical application since poor calibration can lead to overconfidence in incorrect predictions or underconfidence in correct ones, affecting medical decision-making \cite{chua_tackling_2023}. 
This issue is further exacerbated by data variability, class imbalances, and inter-rater disagreements \cite{silva_filho_classifier_2023}. 
Multiple annotation analysis is naturally connected to model calibration and it is meaningful to assume that if multiple raters diverge in their annotations, then a calibrated model should become less confident.

Despite notable progress in uncertainty and calibration research, these aspects are still underrepresented in comprehensive evaluations, often overshadowed by an exclusive focus on predictive accuracy. Critically, the potential of multi-rater annotations to refine calibration metrics remains unexplored. Leveraging such annotations during both training and evaluation could offer a more faithful reflection of clinical ambiguity and improve model robustness. Moving forward, calibration-aware frameworks that explicitly incorporate expert disagreement are essential for building more trustworthy and generalizable medical imaging models.

%\textcolor{red}{
Unlike previous segmentation challenges that primarily focused on label fusion or ignored annotator disagreement during evaluation, CURVAS is the first to systematically benchmark both segmentation accuracy and calibration under multi-rater uncertainty, with a focus on preserving and leveraging inter-rater variability. 
Addressing these obstacles requires rigorous quality control, multi-expert annotations, and robust uncertainty quantification methods. To this end, the Calibration and Uncertainty for multiRater Volume assessment in multiorgan Segmentation challenge (CURVAS) benchmarks algorithm performance on segmentation tasks using abdominal computed tomography (CT) scans with multi-expert annotations. CURVAS, organized as part of the Medical Image Computing and Computer-Assisted Intervention (MICCAI) 2024 conference, provides a structured environment for evaluating segmentation algorithms. Participants used a publicly available dataset\cite{CURVAS_Zenodo} specifically constructed for this competition, featuring annotations from three expert radiologists who segmented three key abdominal organs: the pancreas, kidneys, and liver.%}

%\textcolor{red}{
CURVAS is distinguished not only by its evaluation of segmentation accuracy but also by its emphasis on calibration and uncertainty under multi-rater settings. This paper aims to provide an objective comparison of algorithms through quantitative and qualitative metrics, emphasizing inter-rater variability, calibration, and uncertainty quantification, beyond accuracy. Furthermore, the challenge also assessed the ability of segmentation models to produce meaningful organ volume estimates, a clinically relevant metric, further highlighting each model’s strengths and limitations. A comprehensive analysis reveals that the most accurate models in terms of segmentation are also those best calibrated. Additionally, models trained with pretrained networks or public datasets demonstrated greater robustness to distribution shifts, effectively identifying out-of-distribution regions. These insights contribute to the development of trustworthy AI-assisted medical imaging tools, supporting safer and more effective clinical adoption.%}

\section{The CURVAS Challenge}
\label{sec2}

\subsection{Challenge Organization}
\label{subsec2.1}

The challenge was hosted in the Grand Challenge platform\footnote{https://curvas.grand-challenge.org/} ensuring standardized submission formats, secure data handling, and fair comparison of competing teams.
Participants were invited to train their models from May to end of August 2024.
Subsequently, they were asked to submit their containerized segmentation algorithms to the Grand Challenge website by September 7th 2024. 
Organizers were also permitted to submit containers, which were ranked alongside participant submissions but they were ineligible for prizes. 
The top three submissions were presented by their respective teams during the CURVAS Challenge session on October 10, 2024, at the MICCAI Conference.
The code from the winning team is referenced in the Challenge GitHub repository \cite{curvas2024_github}.

\subsection{Challenge dataset}
\label{subsec2.3}

\subsubsection{Dataset Acquisition and Inclusion Criteria}\label{dataset_acq}
Data collection was approved by the ethics committee at Universitäts-klinikum Erlangen Hospital (approval number 23-243-B) in Bavaria, Germany, and took place between August and October 2023.
In compliance with ethical standards for publishing medical imaging data, both study-specific and broad consent were obtained from all patients, both verbal and written, prior to their participation in the study. 
The data used during the challenge was previously pseudonymized by removing patient-specific and personal information and coded by the hospital. 
This pseudonymization process did not affect the image quality since no metadata related to parameters of the image was affected.

The challenge cohort included 90 CT images prospectively collected.
All participants were over 18 years old, with 51 male and 39 female participants, aged 37–94 years (mean: 65.7 years).
No additional selection criteria were applied to ensure a representative sample of a typical patient cohort.
Inclusion criteria required contrast-enhanced CT scans in the portal venous phase, thin slice acquisition (0.6–1 mm), and a maximum of 10 cysts with diameters under 2.0 cm. 
CT scans with significant artifacts (e.g., breathing artifacts) or incomplete registrations were excluded. 

\begin{figure}[t]
\centering
\includegraphics[width=.49\linewidth]{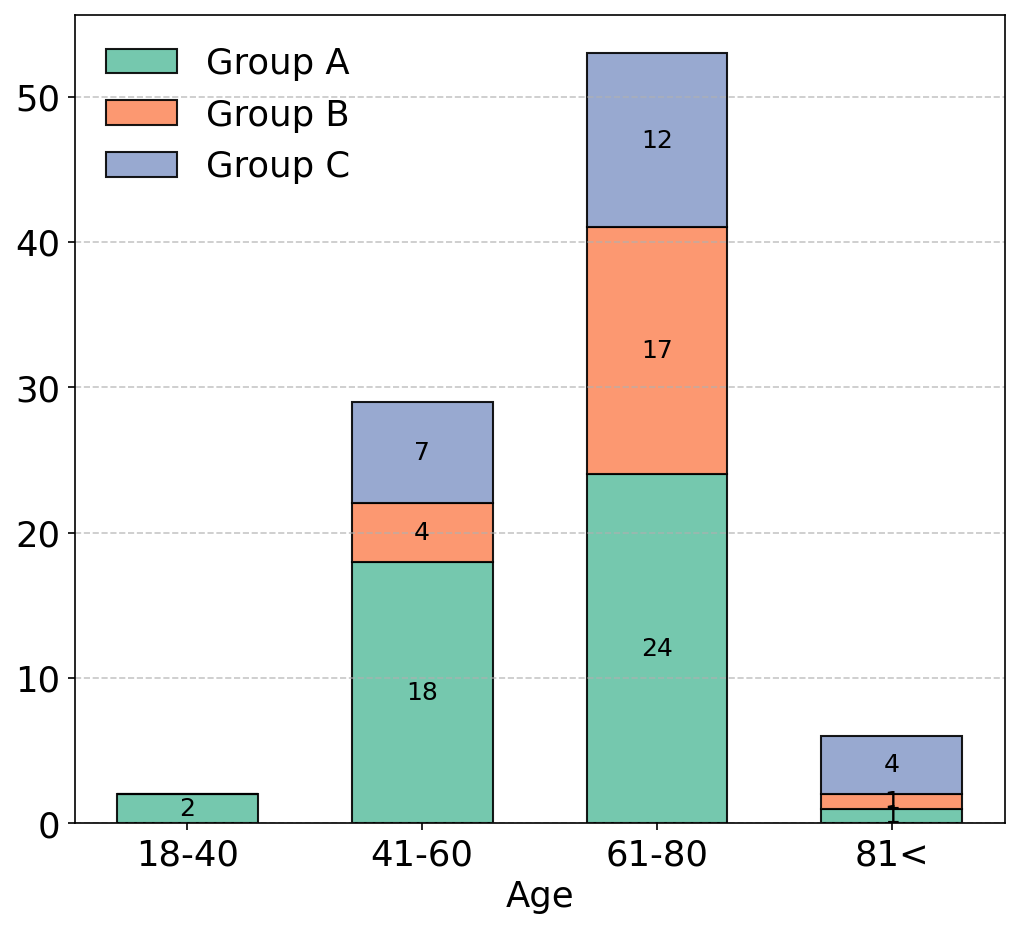}
\includegraphics[width=.49\linewidth]{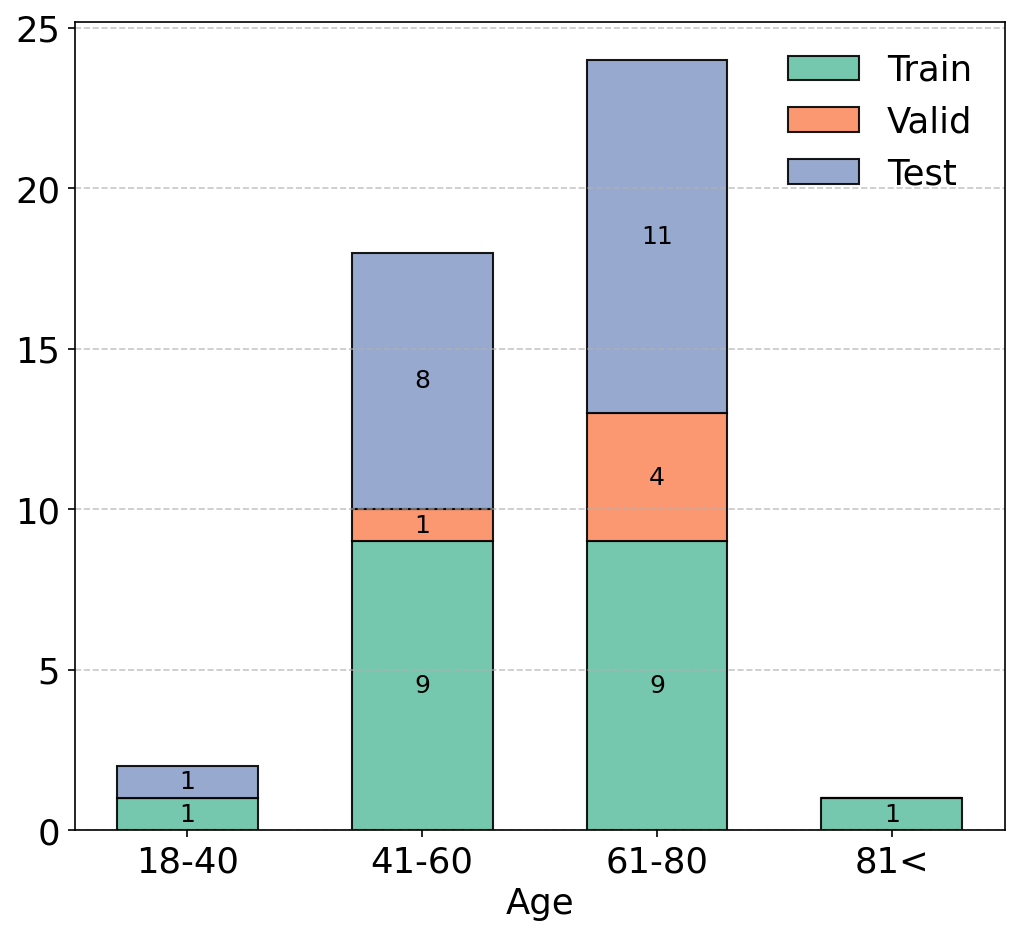}
\caption{Age Distribution Across Groups and Dataset Splits. The left chart shows the distribution of cases across Groups A, B, and C by age range, while the right chart presents the dataset split into training, validation, and test sets. Group B and C cases are more frequent in older individuals, aligning with a higher pathology burden. The dataset split remains balanced, ensuring representation across age groups and pathological complexity.}\label{demography}
\end{figure}

\subsubsection{Data Processing and Annotation Protocol}\label{data_prep}
To prepare the images to be sent to the radiologists, the first step was using Total Segmentator (TS) \cite{totalsegmentator, nnunet} to produce preliminary annotations. 
Then, these annotations, together with their respective CT, were sent to the three radiologists, to both correct the automatic annotations and add possible missing organs that TS had not detected. 
One of the three labeling radiologists, an MD PhD candidate, previously defined both the dataset cohort and the inclusion criteria, as well as the criteria of what belongs to the parenchyma, in order to have common guidelines for all three experts. %, and it was given to the labeling radiologists so they would be coherent with each other. 
Both the dataset cohort with the inclusion criteria and the guidelines were reviewed by another two radiologists, with an experience of 2 and 4 years in abdominal CT imaging.

Each annotation targeted three structures: pancreas, liver and kidneys. 
For the pancreatic organ, it was important the whole pancreas in its course was tracked and marked. 
Neither the splenic vein nor the mesenterial vein were included in the segmentation \cite{westenberger2021automatische}. 
For the kidney structures, both the kidney parenchyma and the renal medulla were included. 
The renal pelvis \cite{brachmann2021evaluation} and the ureter as a urinary stasis, which could alter the original volume, were excluded. 
Lastly, the liver was defined as the entire liver tissue including all internal structures like vessel systems, tumors, etc. \cite{heimann2009comparison}. 
Thus, the portal vein itself and its two main branches were excluded from contouring but any branch of the following generations were included. 
In case of partial enclosure -occurring where large vessels as vena cava and portal vein enter or leave the liver-, the parts enclosed by liver tissue were included in the segmentation, thus forming the convex hull of the liver shape \cite{heimann2009comparison}. 
Any fatty tissue that pulls into the liver was excluded as well as the gallbladder. 
However, wide and especially pathologically widened bile ducts were included in the segmentation of the liver.

\subsubsection{Clinical Subgroups and Data Splitting}\label{data_splits}

Data was separated in three different groups depending on the lesions or pathologies they showed.
Group A (45 CTs) consists on healthier, less complex cases with 2 cysts or less with no contour altering pathologies; Group B (22 CTs) on cases with a moderate presence of cysts (3-5 cysts), providing a slightly more challenging scenario with with no contour altering pathologies; and, Group C (23 CTs) comprises cases with large and more cysts (6-10 cysts) and pathological conditions (liver metastases, hydro nephrosis, adrenal gland metastases, missing kidney), representing the most complex cases for evaluation. 
A visual summary of data distribution is shown Figure \ref{demography}.

The training set consisted on 20 CTs from the Group A.
This limited dataset was intentionally chosen to assess whether the task can be effectively learned with a small number of CT scans or by using publicly available, multi-annotator datasets.
This approach tries to make the participants work in a real world environment where data is limited and tries to make the submitted methods more robust to shifts, allowing participants to develop approaches handling widely accessible data. 
Additionally, training on one dataset while evaluating on another enhances a thorough analysis against distribution shifts and other sources of variability.
For the validation set, only images from Group A (5 CTs) were released, and all the remaining dataset was used for the final testing phase (20 CTs of Group A, 17 CTs of Group B and 23 CTs of Group C).

\subsubsection{Quality assurance}\label{subsubsec2.3.5} 

In the CURVAS Challenge, a rigorous quality assurance (QA) protocol was implemented to ensure the reliability and accuracy of segmentation evaluations. 
Once the images were annotated following the aforementioned protocol, all collected segmentation data underwent thorough integrity verification to confirm the absence of corruption or errors, ensuring that subsequent analyses were based on accurate and consistent data. 
Additionally, to assess annotation consistency, the multiple expert annotations per CT scan were analyzed statistically, focusing on interrater variability to quantify agreement among annotators. 
This process allowed for the identification of potential inconsistencies and provided a deeper understanding of the variability in expert annotations. 
Furthermore, visual inspections were conducted to detect and correct systematic errors or biases in segmentations, ensuring that the annotations were robust across different cases. 

\subsection{Assessment method}
\label{subsec2.4}

\subsubsection{Evaluation metrics}
\label{subsubsec2.4.1}

Multiple metrics were used to evaluate and rank performance. 
To ensure a comprehensive assessment, these metrics were classified into three major groups, each targeting different aspects of model performance. Quality of the Segmentation and Uncertainty Consensus Assessment, Multi-Rater Calibration, and Volume Assessment. 
To account for inter-expert variability, we defined three distinct regions: the foreground consensus area, which is the region that all three clinicians unanimously agreed that the corresponding area belongs to the foreground; the background consensus area which is the region where all three clinicians consistently identified as background; and the dissensus area, which is the region where there was disagreement among the annotators regarding whether it belongs to the foreground or background. 
These predefined regions play a crucial role in the evaluation process, ensuring that model performance is assessed not only in well-defined areas but also in regions with uncertainty, where expert opinions diverge.

\vspace{0.5cm}
\noindent\textit{Quality of the Segmentation and Uncertainty Consensus Evaluation}
\vspace{0.5cm}

The primary segmentation metric used is the Dice Score (DSC), which quantifies spatial overlap between predictions and ground truth. 
DSC evaluation is restricted to the consensus foreground and background regions for three target classes: pancreas, kidney, and liver. 
Consequently, False Positives can only occur in the consensus background area, while False Negatives can only occur in the consensus foreground area.

In addition to segmentation accuracy, we assess uncertainty estimation within consensus regions. 
This analysis is divided into two components: the confidence for the consensus background ($C_{B}$) and the consensus foreground ($C_{F}$) of each organ. 
Subsequently, an overall confidence metric per class ($C_{seg}$) is calculated by integrating both consensus regions, as follows: 

\begin{eqnarray}\label{Cseg}
C_{seg} = \frac{(1-C_{B})+C_{F}}{2} 
\end{eqnarray}

For both DSC and uncertainty assessment, each metric is calculated for each structure individually. 
Then, to obtain the final metrics, the mean across the three organs is computed. 

%%%%%%%%%%%%%%%%%%%%%%%%%%%%%%%%%%%%%%%%%%%%%%%%%%%%%%%%%%%%%%%%%%%%%%%%%%%%
% Second aspect - Calibration
\vspace{0.5cm}
\noindent\textit{Multi Rater Calibration}
\vspace{0.5cm}

For the calibration study, we assess the Confidence Expected Calibration Error ($cECE$), as defined in Equation \ref{ECE}:
\begin{eqnarray}\label{ECE}
cECE = \sum_{m=1}^{M} \frac{|B_{m}|}{M}|acc(B_{m})-conf(B_{m}))|, 
\end{eqnarray}
where $B_{m}$ is the bin with predictive confidence in voxel $m$ is, $acc(B_{m})$ is the accuracy of such bin, $conf(B_{m})$ is the average of such bin, and $M$ is the number of bins.

To preserve multi-rater variability, $cECE$ is computed separately for each prediction against each of the three expert annotations, yielding three distinct $cECE$ values per case. 
To derive a single calibration metric, these three values are averaged equally, ensuring that the annotations of each annotator are considered in the final assessment.

%%%%%%%%%%%%%%%%%%%%%%%%%%%%%%%%%%%%%%%%%%%%%%%%%%%%%%%%%%%%%%%%%%%%%%%%%%%
% Third aspect - Volume
\vspace{0.5cm}
\noindent\textit{Volume Assessment}
\vspace{0.5cm}

For the volume assessment, we incorporate the Continuous Ranked Probability Score (CRPS) to evaluate how well the predicted volumetric distributions align with the ground truth. 
Unlike other metrics, CRPS provides a more clinically relevant measure by considering the full probabilistic distribution of the predictions.
To retain multi-annotator variability, we define a Gaussian Probability Distribution Function (PDF) based on the mean and standard deviation of the volumes derived from the three expert annotations. 
From this, we compute the corresponding Cumulative Distribution Function (CDF).

\begin{figure}[t]
\centering
\includegraphics[width=\linewidth]{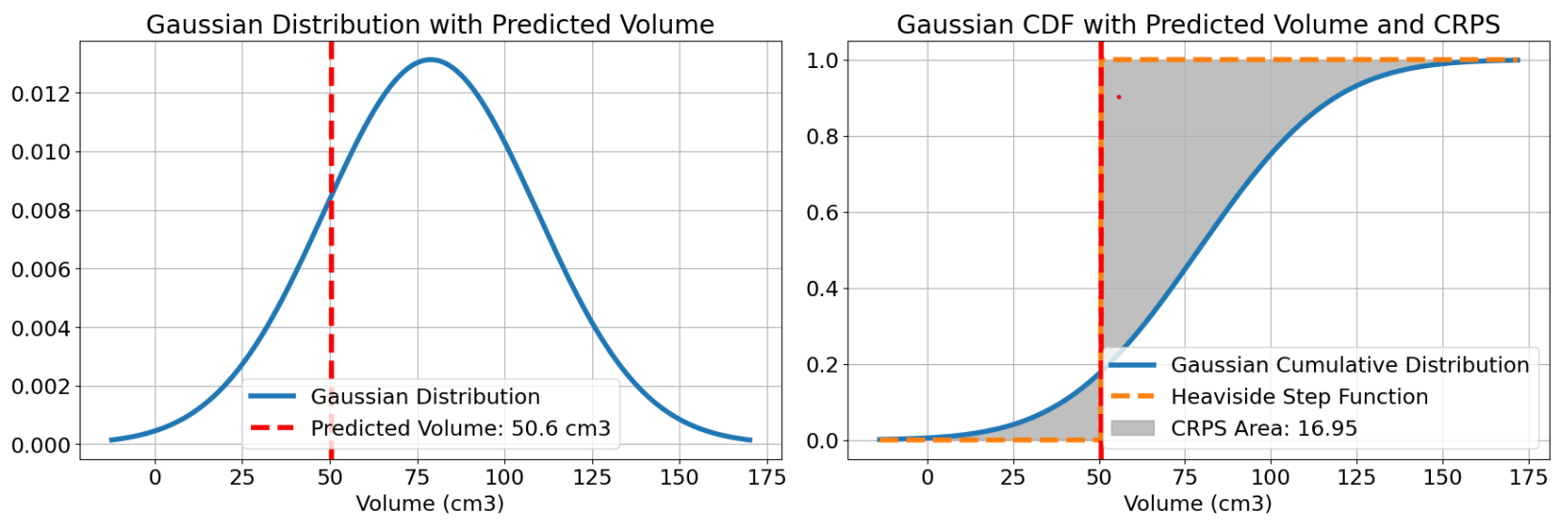}
\caption{Visual example of a CRPS calculation. \textit{Left figure}: Gaussian Probability Distribution function (PDF) \textit{blue line} with the predicted volume (\textit{red line}). \textit{Right figure}: Gaussian Cumulative Distribution function (CDF) (\textit{blue line}) with the predicted volume (\textit{red line}) and its corresponding Heavyside representation and the CRPS area.}\label{crps_example}
\end{figure}

The predicted volume is obtained by summing all probabilistic values for the corresponding class from the probabilistic output provided by the participant. 
This approach integrates the model's uncertainty into the volume estimation. 
Equation (\ref{CRPS}) represents the Continuous Ranked Probability Score (CRPS), computed as the average squared difference between the cumulative distribution and the predicted value. 

\begin{eqnarray}\label{CRPS}
CRPS(F,y)=\int(F(x)-1_{\{x\geq y \}})^{2}dx,
\end{eqnarray} 
$F(x)$ being the PDF obtained from the ground truths and $1_{\{x\geq y \}}$ being the Heaviside function of the volume calculated from the prediction. 
For a graphical representation of this equation see Figure \ref{crps_example}.

For a clearer understanding, Figure \ref{crps_example} illustrated the probabilistic volume evaluation process. 
The left panel displays the Gaussian PDF (\textit{blue line}), representing the volume distribution of a structure, computed from the mean and standard deviation of the three expert annotations. 
The red line indicates the predicted volume by the model. 
The right panel presents the corresponding CDF (\textit{blue line}) derived from the Gaussian PDF, with the red line again representing the predicted volume. 
Additionally, the orange line corresponds to a Heaviside function derived from the predicted volume, while the gray area quantifies the CRPS computation. 
A smaller gray area signifies a predicted volume closer to the ground truth distribution, indicating better performance.

In conclusion, by using the CRPS, we evaluate the accuracy of the probabilistic volume prediction in comparison to the ground truth volume's probabilistic distribution. 
The CRPS is a useful metric for assessing how well the predicted CDF aligns with the true CDF of the volume, accounting for both the prediction's central tendency and its spread. 
This approach allows us to quantify the difference between the predicted and actual distributions over the entire volume range, providing a comprehensive measure of prediction quality. 
Minimal CRPS values reflect improved model reliability and confidence in the accuracy of the estimated volume distribution.

\subsubsection{Ranking}
\label{subsubsec2.4.2}

Each participating team was ranked according to four evaluation metrics: DSC, $C_{seg}$, ECE, and CRPS. 
There were four separate rankings: two in descending order (DSC and $C_{seg}$) and two in ascending order (ECE and CRPS). 
The final ranking was determined by combining these four individual rankings. 
For each algorithm, the relative ranks across all metrics were averaged, producing a composite score. 
The algorithms were then ranked based on their average relative rank, with a lower mean indicating better overall performance. 
This approach ensures fairness by placing all metrics on a comparable scale, allowing the final ranking to reflect balanced performance across all evaluation criteria, without bias towards any specific metric.

\subsubsection{Further analysis}
\label{subsubsec2.4.3}

In addition to evaluating the performance of different algorithms, this paper presents a comprehensive analysis of their results. 
First, we study the general ranking of the algorithms based on overall performance across multiple metrics, providing an aggregated view of their effectiveness. 
In addition, we identify the best performing algorithm for each specific metric, providing insight into strengths and weaknesses across different evaluation criteria. 
This detailed breakdown allows for a nuanced understanding of how each method excels or struggles depending on the metric under consideration.

Beyond overall rankings, we analyze the variability of performance both per algorithm and per metric, highlighting the consistency or instability of each method. 
To further explore these variations, we conduct a qualitative analysis of metric distributions at the image level, examining cases where algorithms perform particularly well or poorly with a visual example. 
Additionally, we investigate performance across the different groups $(A, B, C)$, revealing trends and discrepancies that may be associated with specific data characteristics. 
Finally, we assess correlations between different metrics, identifying potential dependencies and redundancies that can refine future evaluation protocols. 
This multi-faceted analysis aims to provide a deeper understanding of algorithm behavior, ultimately guiding the development of more robust and reliable models.

Finally, a thorough statistical analysis is conducted to evaluate the stability of the rankings across all the methods. 
This analysis involves performing bootstrap sampling, a technique that allows us to estimate the variability of the rankings by repeatedly sampling from the original dataset with replacement. 
By evaluating how the rankings fluctuate across multiple bootstrap iterations, we can assess the consistency of the methods’ performance. 
This approach helps to determine whether the observed differences in rankings are robust or if they are likely due to sampling variability.

\section{Team Methods}\label{sec3}

The challenge included seven participating teams from various institutions. The MedIG team from the Shenzhen Institute of Advanced Technology, Chinese Academy of Sciences, and the University of Chinese Academy of Sciences (China); the PrAEcision team from Friedrich-Alexander-Universität Erlangen-Nürnberg (FAU) (Germany); the BreizhSeg team from Université de Rennes 1, CLCC Eugène Marquis, and INSERM UMR 1099 LTSI (France); the DLAI team from Imperial College London and University College London (United Kingdom); the BCNAIM team from Universitat de Barcelona and ICREA (Spain); the CAI4CAI team from King's College London (United Kingdom); and the PredictED team from Universitat de Barcelona (Spain) and the University of Edinburgh (United Kingdom). Each team applied distinct strategies and methodologies, which are further detailed in the following subsections.

\subsection{MedIG}\label{sec3.1}

To address variations among labels provided by multiple annotators, the STAPLE algorithm \cite{warfield2004staple} was used to generate consensus labels for the pancreas, kidney, liver, and background. Applied independently to each organ and background, this approach produces robust consensus segmentations that integrate annotations across all annotators.

The 3D full-resolution version of nnUNet \cite{nnunet} was selected, a state-of-the-art deep learning-based method for medical image segmentation that leverages 3D spatial information in CT images. The network employs a vanilla U-Net architecture with six downsampling steps and a decoding phase that integrates skip connections and upsampled feature maps. To meet the time constraints of the Grand Challenge platform, MedIG implemented post-processing to accelerate probability map handling by setting values below $1 \times 10^{-6}$ to zero, reducing storage demands and optimizing compression. This approach minimized I/O operations, ensured efficient data handling, and allowed the algorithm to execute swiftly within the platform's requirements.

\subsection{PrAEcision}\label{sec3.2}

Given the promising results of the nnUNet model \cite{nnunet} for general medical imaging segmentation tasks \cite{huang2024segmentingmedicalimagesunet} and specifically on multi-rater annotation segmentation challenges \cite{li2024qubiquncertaintyquantificationbiomedical}, PrAEcision trained a 3D lowres nnUNet model for the CURVAS multirater segmentation challenge to segment the organs background, kidney and liver. Further, they sought to determine whether the practical training and configuration of nnUNets effectively translate to real-world inference challenges, including resource and deployment constraints. To encompass the variations in segmentations by different annotators, not only the available abdominal CT scans from Uniklinikum Erlangen provided on the challenge website were used, but also the publicly available abdominal CT datasets were included in the model training process. Specifically, the datasets WORD \cite{WORD}, Amos \cite{AMOS}, TotalSegmentator (TS) \cite{totalsegmentator}, AbdomenCT-1K \cite{9497733} were used. 

Due to the inference time limitations, the 3D lowres model was submitted as its inference was faster than both the 3D fullres and 2D nnUNet model and any ensemble model. Although the 2D model delivered the best performance on the validation set in terms of DSC, the 3D lowres model was chosen to meet the inference time constraints.

\subsection{BreizhSeg}\label{sec3.3}

To address the fact that the training dataset is limited, TS \cite{totalsegmentator} was leveraged and incorporated uncertainty quantification using Adaptable Bayesian Neural Network (ABNN) \cite{franchi2023makebnnsimplestrategy}. ABNN transforms this deterministic network into a Bayesian one by introducing Bayesian Normalization Layers (BNL), which replace standard layer normalization. In BNL, uncertainty is modeled by sampling the trainable normalization parameters, $\gamma$ and $\beta$, whose values are empirically estimated during fine-tuning. Gaussian perturbation ($\epsilon$) is applied to $\gamma$ before training, injecting randomness that enables the model to approximate the posterior distribution of the weights. In this approach, $\epsilon$ is sampled from a reduced centered Gaussian distribution with reduced variance. To capture uncertainty, $\gamma$ and $\beta$ are obtained by retraining the model 20 times for 10 epochs each, while keeping all other weights fixed. This fine-tuning strategy focuses exclusively on learning variability in the normalization parameters. A five-fold cross-validation was conducted to ensure dataset diversity. For each fold, the model was retrained under four distinct annotation settings: three runs using annotations from individual physicians and one run using the combined annotations from all three.

Training adhered to TS’s constraints, optimization scheme, and loss function, with an exponentially decaying scheduler for efficient convergence. Class weights for target structures varied across epochs to enhance uncertainty estimation, while weights for 22 additional anatomical labels were fixed, extending the model’s segmentation scope. During inference, the Bayesian model estimates uncertainty through multiple predictions. Preprocessing includes resampling to a consistent voxel size, standardizing orientation, and segmenting images into overlapping 3D patches. Patch-wise predictions are combined using a cosine function for spatial coherence, and uncertainty maps are generated by averaging prediction scores, producing confidence values between 0 and 1. Post-processing retains the largest connected volumes in segmentation and uncertainty maps, ensuring anatomically meaningful results. Preprocessing steps are then reversed to return CT images and segmentation maps to their original format for evaluation and interpretation.

\subsection{DLAI}\label{sec3.4}

Convolutional Neural Networks (CNNs) and Vision Transformers (ViTs) have driven advancements in biomedical image segmentation but struggle with capturing long-range dependencies due to locality and computational constraints. To overcome these limitations, DLAI proposes xLSTM-UNet, a UNet-based deep learning model enhanced with an efficient parallel cross-window module. This architecture integrates Vision-LSTM (xLSTM) as its backbone, leveraging xLSTM's ability to capture long-range dependencies effectively, as demonstrated in Vision-LSTM applications for NLP and image classification \cite{alkin2024visionlstmxlstmgenericvision}. By combining xLSTM’s strengths with UNet’s local feature extraction, xLSTM-UNet delivers comprehensive image analysis capabilities \cite{chen2024xlstmuneteffective2d}.

To further enhance segmentation performance, a Cross Attention Transformer (CAT) block into the encoder is incorporated. 
This block uses Window-based Multi-head Cross Attention (W-MCA) \cite{QAYYUM2024102226} to compute feature representations by cross-attending between base and searching windows. 
Updated features are processed through a two-layer MLP with GELU activation, and LayerNorm (LN) ensures stability. 
This enables efficient fusion of features across image regions, improving segmentation accuracy.

This model was implemented in PyTorch and trained with the Adam optimizer for 500 epochs. The loss function was a combination of Cross-Entropy and Dice Loss for optimization and used a sliding window technique for inference. Data preprocessing included resampling, z-score normalization, and region-focused cropping. Data augmentation strategies included spatial transformations (e.g., rotations, scaling, mirroring), intensity adjustments (e.g., Gaussian noise, brightness, contrast), gamma corrections, and elastic deformations to simulate anatomical variability. These steps ensure robust learning and generalization across diverse datasets. For the inference, the sliding window technique was used.

\subsection{BCNAIM}\label{sec3.5}

Public abdominal CT datasets provide a valuable resource for pretraining 3D segmentation networks like UNet3D \cite{10.1007/978-3-319-46723-8_49} or SwinUNETR \cite{9879123}, which can then be fine-tuned for specific challenges. However, variations in the number of annotated organs across datasets complicate their combined use. Liu et al. \cite{Liu_2023} addressed this by proposing a universal model with CLIP-driven conditioning that dynamically generates segmentation head parameters based on the target organs. The model combines visual features from a backbone (e.g., UNet or SwinUNETR) with text embeddings generated from a prompt, “A CT of a [CLS],” where [CLS] corresponds to the target organ. This hybrid approach facilitates the use of diverse datasets.  

Building on this, the model used pretrained weights provided in the AbdomenAtlas 1.0 repository \cite{abdomen_atlas_github}, providing the largest CT dataset to date \cite{qu2023abdomenatlas8kannotating8000ct, LI2024103285, li2024how, bassi2024touchstone}. Given its smaller number of parameters and faster training time, the UNet backbone was selected for fine-tuning on the CURVAS Challenge dataset. Fine-tuning the entire vision pathway proved more effective than tuning only the controller, despite increased training time. To handle annotation protocol differences, three models were trained starting from AbdomenAtlas-8K weights, and their logits were ensembled and passed through a sigmoid function to generate final segmentation masks.  

Preprocessing steps included downsampling CT images to a 1.5 mm$^{3}$ resolution, intensity clipping to [-175, 250], and cropping to reduce air outside the patient. Fine-tuning involved extracting 3D patches $(96\times96\times 96)$ from regions with non-zero ground truth segmentation, and testing used sliding window patches with 50\% overlap. To optimize inference time, images were preprocessed, and an initial segmentation mask was generated with one fine-tuned model before cropping the area of interest. This reduced region was then processed by all three models to produce the final segmentation mask.

\subsection{CAI4CAI}\label{sec3.6}

The submitted algorithm was trained using the MONAI framework with a SegResNet model from scratch, utilizing all 20 training images. The loss function combined Dice and Cross-Entropy with an auxiliary soft-binned Average Calibration Error loss \cite{barfoot_average_2024}, where soft-binning assigned probabilities to bins based on their linear distance rather than a hard assignment. The loss components were weighted equally. Training employed the Adam optimizer with a learning rate of 0.0001, using a WarmupCosineSchedule over 200 epochs. Multiple annotations were treated as separate samples, and patch-based training was applied with $144\times144\times144$ patches, with images resampled to 2.00mm³ due to memory constraints. Data augmentation was minimal, limited to intensity scaling and random affine transformations.

\subsection{PedictED}\label{sec3.7}

The ensemble model consists of two U-Net++ architectures \cite{8932614}, with the first using an EfficientNet-B3 \cite{pmlrv97tan19a} encoder and the second employing a MobileViTv2\_050 encoder \cite{mehta2022mobilevitlightweightgeneralpurposemobilefriendly}, both pretrained on ImageNet. Each model processed 512x512 single CT slices, replicated across three channels, and produces a four-channel softmax output (background plus three organs). During inference, predictions from both models were combined through weighted averaging, with Model 2 receiving greater weight due to its ability to capture finer details in the kidney and liver regions.

For training, focal loss \cite{8237586} was used to address class imbalance, particularly for the pancreas, while AdamW optimizer \cite{loshchilov2019decoupledweightdecayregularization} and OneCycleLR scheduling \cite{smith2018disciplinedapproachneuralnetwork} were employed. The model was trained for 100 epochs with a batch size of 64, using data augmentation techniques such as rotations, flips, shifts, scaling, and brightness/contrast adjustments. Mixed precision training is applied to enhance convergence speed and reduce memory usage. A content-aware sampling strategy ensures a balanced ratio of empty and non-empty slices per batch. During inference, test-time augmentation (TTA) is applied by averaging predictions from the original and horizontally flipped slice, improving stability and robustness to geometric variations \cite{perez2017effectivenessdataaugmentationimage}.

\begin{landscape}
\begin{table}[ht]
\centering
\fontsize{8}{9.5}\selectfont  %
\begin{adjustbox}{width=\linewidth, totalheight=0.96\textheight}
\begin{tabular}{p{1.2cm} p{1.6cm} p{1.8cm} p{1.7cm} p{2cm} p{2.2cm} p{1.8cm} p{2.2cm} p{1.6cm}}
  \textbf{Algorithm} & \textbf{Annotations} & \textbf{Network} & \textbf{PatchSize*} & \textbf{DA} & \textbf{Loss Function} & \textbf{Optimizer**} &  \textbf{Public Data} & \textbf{GPU***}\\
  \toprule
  MedIG & STAPLE & nnUNet 3D full-res with ResidualEncoderUNet & $256\!\times\!160\!\times\!160$ \newline $1.0,0.96,0.96$ & Default 3D augmentation params (mirror augmentation disabled) in nnUNet & CE + DSC & SGD$^{1}$ \newline 0.01 with Polynomial Decay  & No & S NVIDIA GeForce RTX 3090 \\
  
  PrAEcision & All annotations used separately & nnUNet 3D low-res & $96\!\times\!160\!\times\!160$ \newline $2.48,1.61,1.61$ & spatial augmentation techniques & CE + DSC & SGD$^{1}$ \newline 0.01 with Polynomial Decay & WORD, Amos, TS, AbdomenCT-1K & A100 GPU\\
  
  BreizhSeg & TS fine-tuned 20 times using four expert-specific training sets and one combined set$^3$ & Adaptable Bayesian NN \& TS & $128\!\times\!128\!\times\!128$ \newline $1.5,1.5,1.5$ & - & CE + DSC with adaptive weighting (0.5–1 for targets, 0.2 for others) & AdamW \newline 0.0001 with Exponential Decay & pretrained TS & NVIDIA RTX A6000 with 48GB VRAM\\
  
  DLAI & All annotations used separately & xLSTM-UNet adding a CAT* block within the encoder & $128\!\times\!128\!\times\!128$ \newline $1,1,1$ & Spatial (rotations, scaling, mirroring) and intensity (noise, brightness, contrast, gamma) augmentations, plus elastic deformations. & CE + DSC & Adam \newline 0.0001 & No & Tesla A6000 GPU with 48GB \\
  
  BCNAIM & Model per annotator & UNet & $96\!\times\!96\!\times\!96$ \newline $1.0,0.96,0.96$ & MONAI transforms$^{2}$ & CE + DSC & AdamW \newline 0.0001$^4$ & AbdomenAtlas-pretrained weights & NVIDIA GeForce RTX 4090 with 24GB\\ 
  
  CAI4CAI & All annotations used separately & MONAI: SegResNet & $144\!\times\!144\!\times\!144$ \newline $2,2,2$ & Scale intensities and random affine transforms & CE + DSC + Average Calibration Error (weighted equally) & Adam \newline 0.0001 with WarmupCosineSchedule & No & - \\ 
  
  PredictED & Majority Voting & Ensemble of two models: UNet++ with two encoders (EfficientNet-B3 and Mobilevitv2\textunderscore050) & $512\!\times\!512$ (2D) \newline - & Random rotations ($90°$), flips, shift, scale, rotate, brightness and contrast adjustments & Focal Loss & AdamW \newline Scheduler: OneCycleLR & ImageNet-pretrained encoders &  -
\end{tabular}
\end{adjustbox}
\caption{Overview of algorithms submitted to the CURVAS Challenge ordered from best to worst.}\label{tab:algorithms}
{\tiny
* The Spacing is specified below the Patch Size
** The Learning Rate is specified below the Optimizer
*** GPU used for training, for inferences all teams had the same environments
(1) with momentum of 0.99 \& weight decay of 0.001
(2) RandCropByPosNegLabel crops fixed-size regions centered on a foreground or background voxel (pos=2, neg=1); RandRotate90 applies random 90° rotations (max k=3), and RandShiftIntensity shifts intensity within a 0.10 range.
(3) each set followed a five-fold cross-validation strategy (4 sets × 5 folds = 20 runs)
(4) with the weight decay $1\times 10^{-5}$ \& Linear warmup from $lr=0.0$ to 0.0001 for 100 epochs, then cosine annealing decay for the rest of the training
}
\end{table}
\end{landscape}

\subsection{Teams' summary}
\label{subsec3.8}

Out of the seven teams, three (MedIG, BreizhSeg, PredictED) submitted  two algorithms, meaning the total number of submissions was ten.
The different teams' algorithms are summarized in Table \ref{tab:algorithms}.
All of the submissions were based on a UNet architecture, each with its own variations.
Two of the teams, MedIG and PredictED, incorporated merging techniques for the manual annotations provided (using STAPLE and Majority Voting, respectively), while the others opted not to use any merging and instead trained on all labels.
All participants trained with small 3D patches, and during inference, they employed the sliding window technique with specific adaptations for each model, except for one, which trained using 2D images from entire CT slices.
%\textcolor{red}{
Preprocessing strategies varied: MedIG, PrAEcision, and BCNAIM standardized voxel spacing and used intensity normalization; BCNAIM also performed cropping centered on organ regions, while CAI4CAI scaled intensities and applied affine transformations; PredictED worked on native 2D CT slices without resampling. Augmentation usage ranged from none (BreizhSeg) to highly diverse pipelines: DLAI combined spatial (rotations, scaling, mirroring) and intensity (noise, brightness, contrast, gamma) transformations with elastic deformations; MedIG used nnU-Net’s default 3D augmentations with mirroring disabled; PrAEcision applied spatial transformations; BCNAIM used MONAI transforms targeting both spatial and intensity variability; PredictED employed 2D geometric (rotations, flips, shifts, scales) and photometric (brightness, contrast) augmentations.
%}
Most algorithms combined Dice Loss and Cross Entropy for training.
However, PredictED used Focal Loss to address class imbalance and CAI4CAI used a combination of Dice Loss and Cross Entropy, as other teams, but added a calibration loss based on Average Calibration Error.
Lastly, four teams (PrAEcision, BreizhSeg, BCNAIM, and PredictED) incorporated external knowledge, either through pretrained models or public external databases.

\section{Results}
\label{sec4}

\subsection{Quantitative results}
\label{subsec4.3}

In this section, we analyze how the adopted metrics provide complementary insights and supplement each other. Next, we examine the overall results of the challenge, and finally, we evaluate these metrics for each group.

\subsubsection{Metrics Analysis}

\begin{figure}
    \centering
    \includegraphics[width=\linewidth]{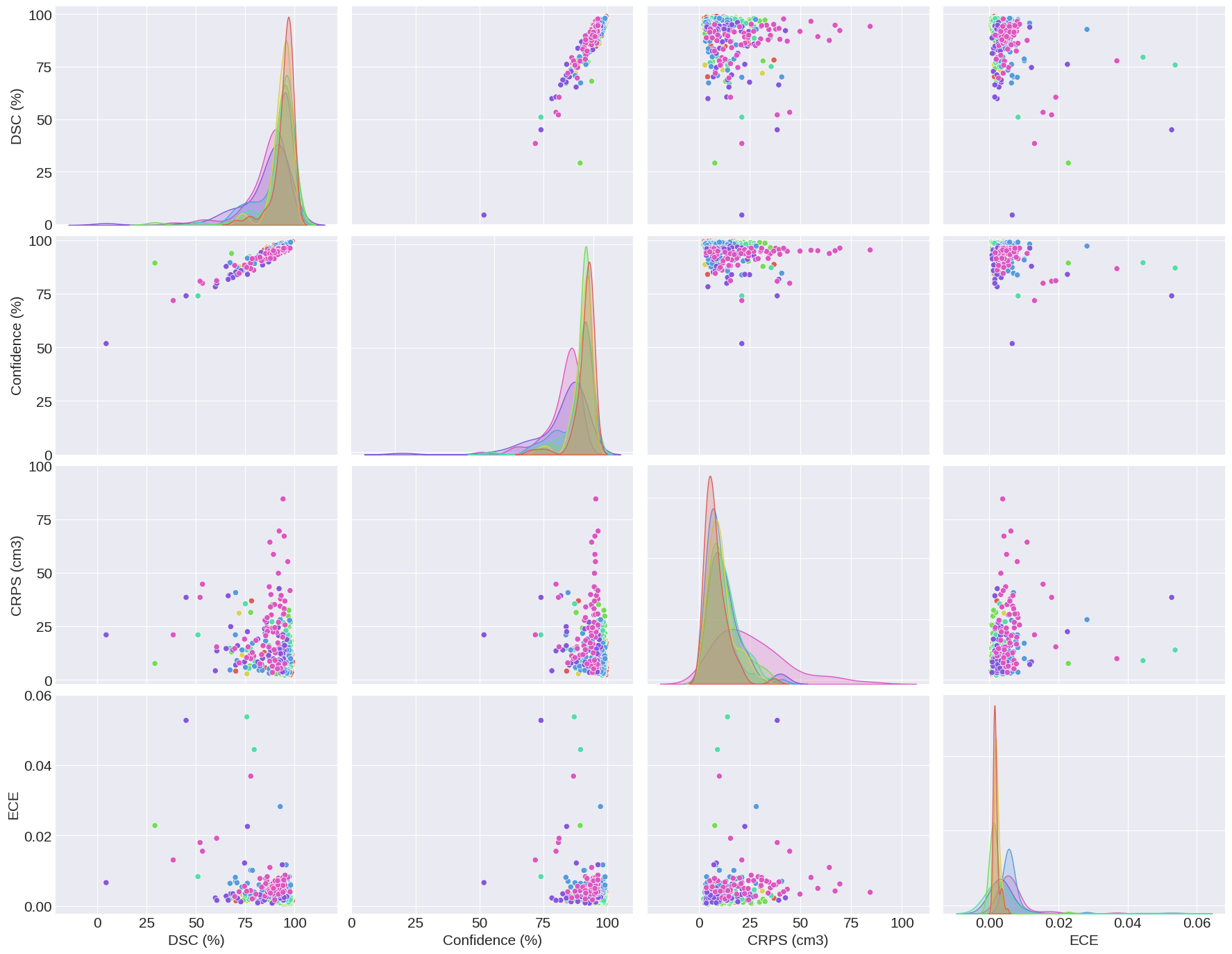}
\caption{Pairwise comparison of the different metrics across all algorithms. The diagonal shows the distributions of each metric, while the off-diagonal plots depict the relationships between pairs of metrics. Each color represents a specific algorithm, consistent with the color scheme used in the boxplot Figure \ref{fig:metrics}, ensuring direct comparability between the two visualizations.}
\label{fig:correlation}
\end{figure}

Figure \ref{fig:correlation} shows an analysis across participants of the relationship between the different considered metrics.
We first focus on the diagonal plots, which represent the distribution of each metric across all algorithms. 
For DSC and Confidence, the distributions are heavily skewed toward the higher end (close to 100\%), indicating that most predictions exhibit high segmentation accuracy and confidence, which is a reasonable outcome considering we are working with relatively large and stable anatomical structures. 
CRPS, on the other hand, shows a wider range, with most values clustering below 20 cm$^3$  and a long tail. 
Meanwhile, ECE values are relatively low, reflecting strong calibration across all algorithms.

The off-diagonal subplots in Figure \ref{fig:correlation} allow us to examine correlations between metrics and reveal several trends. 
A clear positive correlation is observed between Confidence and DSC, with higher confidence generally corresponding to higher DSC values, although some outliers exist. 
Similarly, lower CRPS values are associated with higher DSC scores. %, as expected, since lower CRPS indicates better probabilistic predictions. 
For Confidence versus CRPS, a cluster is evident at high confidence and low CRPS values, but variability increases at lower confidence levels. Lastly, no strong relationship is apparent between ECE and DSC, as ECE values remain close to zero regardless of DSC. Similarly, Confidence and ECE exhibit only minimal correlation, with ECE remaining consistently low across the range of Confidence values.

It is important to emphasize that the four considered metrics complement each other well, as they are not highly correlated, providing unique insights into algorithm performance. 
The strongest relationship is between DSC and Confidence, which aligns with expectations given the anatomically consistent structures to segment in this challenge, except for the pancreatic parenchyma. 
For more complex or smaller structures, such as tumors, distinguishing between DSC and confidence could prove more valuable. 
High Confidence in both positive and negative segmentations could play a critical role in guiding future follow-ups or decisions for intervention.

\subsubsection{Overall Algorithm Performance} % Global Performance Assessment

Table \ref{tab:metrics} presents the aggregated metrics, offering a clear summary of each submission's performance.
The results highlight that the top three teams in both Confidence and ECE metrics remain consistent, even though for the overall calibration of the model, BreizhSeg has the best score. 
For the DSC metric, DLAI slightly outperforms BreizhSeg, but the difference is minimal ($\sim 0.12\%$).
Interestingly, in the CRPS metric, BCNAIM achieves a second-place ranking. 
This suggests that while their confidence may be lower in consensus areas, their predictions align more closely with the ground truth overall, showing high probabilities in positive regions and near-zero probabilities elsewhere.
However, this contrasts with their 6th-place ranking in ECE, indicating inconsistencies in calibration despite strong CRPS performance.
CAI4CAI and PredictED have the lowest dices.
This is likely because PredictED used 2D slices for training, limiting its ability to capture fine segmentation details and spatial information.
Additionally, CAI4CAI's training minimized a loss that incorporated both DSC and calibration as the minimization goal. 
This is further supported by CAI4CAI's superior ranking in the ECE metric compared to the other four methods.

\begin{table}[t]
\centering
\fontsize{11}{11}\selectfont
\begin{tabular}{c|c|c|c|c}
  \textbf{Algorithm} & \textbf{DSC (\%)} & \textbf{Confidence (\%)} & \textbf{ECE ($\times 10^{-3}$)} & \textbf{CRPS ($cm^{3}$)}\\
  \hline
  \textbf{MedIG} & \textbf{94.57 (1)} & \textbf{97.87 (1)} & \textbf{1.82 (2)} & \textbf{8.108 (1)} \\ %8108.4840
  \textbf{PrAEcision} & \textbf{93.29 (2)} & \textbf{97.18 (2)} & \textbf{2.22 (3)} & \textbf{10.438 (3)} \\ %10438.1709
  \textbf{BreizhSeg} & 92.60 (4) & \textbf{97.17 (3)} & \textbf{1.61 (1)} & 12.326 (4) \\ % 12325.7742
  \textbf{DLAI} & \textbf{92.72 (3)} & 96.23 (4) & 3.90 (4) & 12.625 (5) \\ %12625.4213
  \textbf{BCNAIM} & 90.52 (5) & 95.88 (5) & 6.21 (6) & \textbf{9.727 (2)} \\ %9727.3923
  \textbf{CAI4CAI} & 84.98 (7) & 92.10 (7) & 4.48 (5) & 12.828 (6) \\ %12828.0093
  \textbf{PredictED} & 85.79 (6) & 92.39 (6) & 6.64 (7) & 25.895 (7) \\ % 25895.3869
\end{tabular}
\caption{Comparison of the different teams' metrics, ordered according to their final ranking from best to worst. The relative rankings for each metric are indicated in parentheses in each columns respectively.}\label{tab:metrics}
\end{table}

\begin{figure}
    \centering
    \includegraphics[width=.49\linewidth]{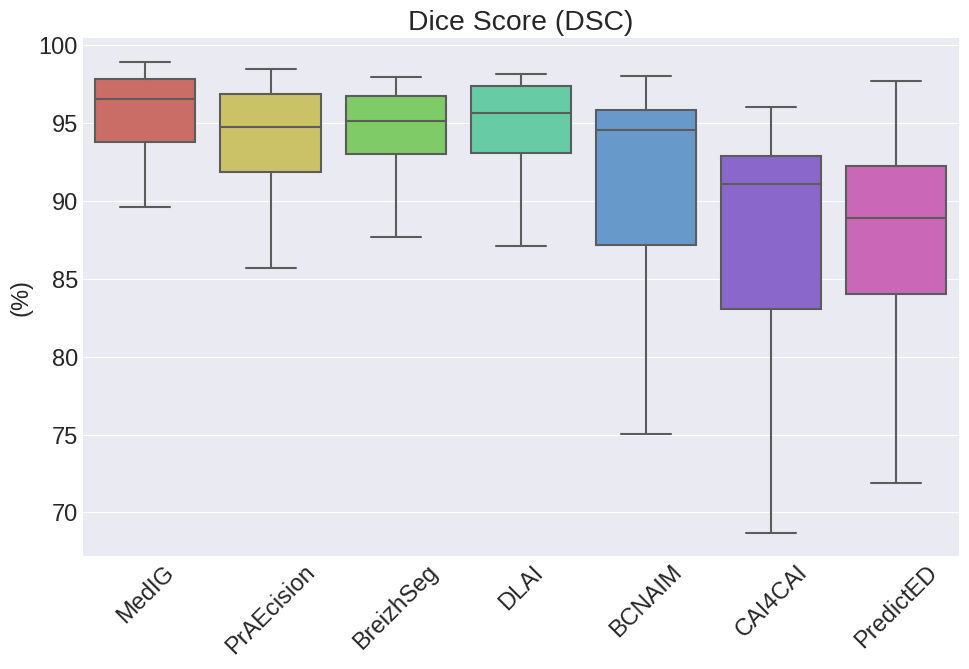}
    \includegraphics[width=.49\linewidth]{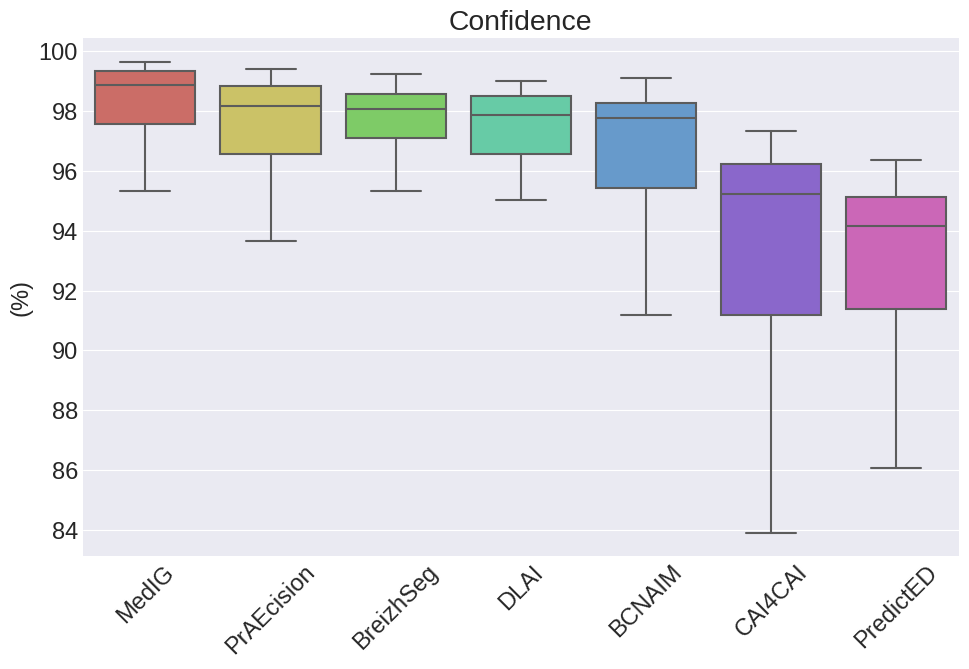}
    \\[\smallskipamount]
    \includegraphics[width=.49\linewidth]{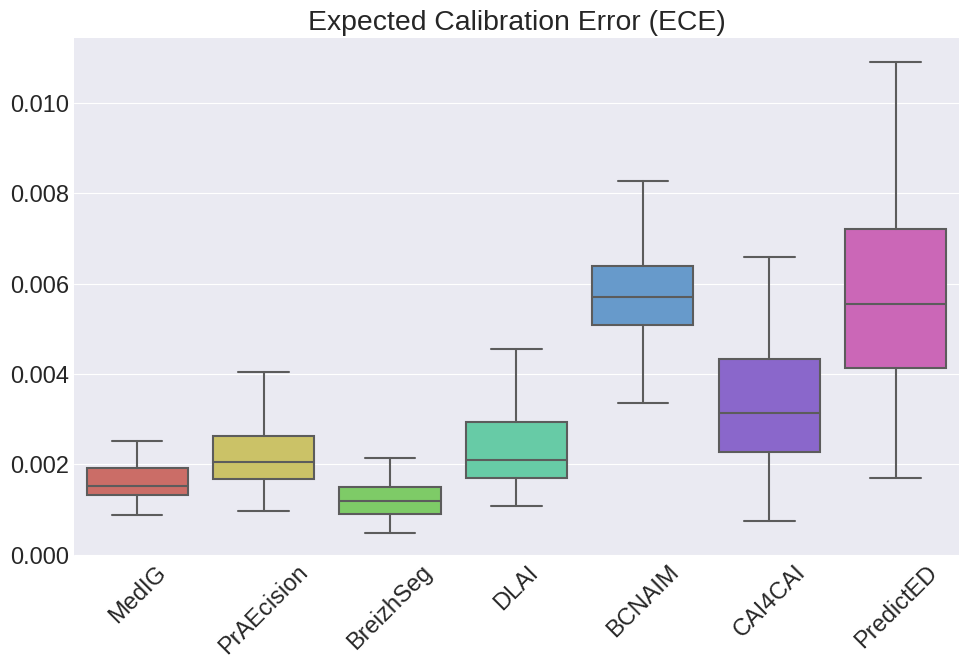}
    \includegraphics[width=.49\linewidth]{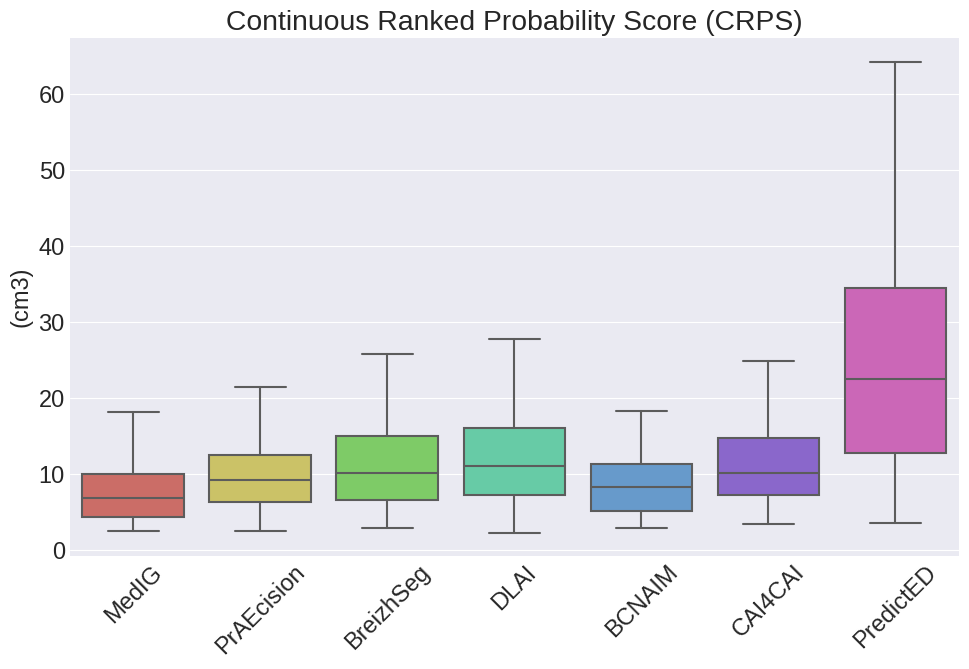}
\caption{Boxplots of the evaluation metrics (DSC, Confidence, ECE, and CRPS) for all algorithms. Each box represents the distribution of metric values for a specific algorithm, with each color corresponding to a specific algorithm. In this plot the outliers are not considered. Teams ranked from best to worst in the final ranking are visualized from left to right on the x-axis of each plot.}
\label{fig:metrics}
\end{figure}

Figure \ref{fig:metrics} further illustrates these findings, showing that algorithms with better overall performance tend to exhibit less variation, reflecting greater stability and robustness. For example, in terms of the ECE metric, BreizhSeg not only achieves the best calibration but also displays minimal variation. The plot reinforces the trends observed in Table \ref{tab:metrics}, emphasizing the relationship between high performance and stability across metrics.

\subsubsection{Analysis Per Clinical Group}

In our per-group analysis, we would like to ascertain if there are systematic performances between group A (healthier, less complex cases with 2 cysts or less with no contour altering pathologies), group B (moderate presence of 3-5 cysts, providing a slightly more challenging scenario with with no contour altering pathologies), and group C (cases with large 6-10 cysts and pathological conditions -liver metastases, hydro nephrosis, adrenal gland metastases, missing kidney-, representing the most complex cases). See Subsection \ref{data_splits}.

Table \ref{tab:groupwise_metrics1} and \ref{tab:groupwise_metrics2} present the performance metrics for each algorithm across the different clinical groups. MedIG consistently achieves the best overall performance, except for ECE, where BreizhSeg outperforms all other methods in both Group A and Group C. PrAEcision and BreizhSeg remain close competitors in most metrics, though MedIG maintains a clear advantage in CRPS, indicating superior volume estimation accuracy and a more precise calibration in regression, as reflected by more reliable probability estimates.

The focus of group-wise performance analysis, however, is not just the absolute performance of each algorithm but the trends observed across groups and whether these trends remain consistent for all methods. A clear pattern emerges: segmentation quality (measured by DSC) decreases progressively from Group B to Group A, reaching its lowest values in Group C. This aligns with previous observations that Group C presents the greatest challenges, containing more complex anatomical structures and ambiguous cases. Similarly, volume estimation errors (CRPS) increase in the same order (Group B $\xrightarrow{}$ Group A $\xrightarrow{}$ Group C) mirroring the trend in DSC values.

Regarding calibration, BreizhSeg achieves the best ECE values in Groups A and B, while MedIG surpasses it in Group C, suggesting that these models produce probability estimates that are more aligned with actual segmentation correctness. In contrast, DLAI and PredictED exhibit poor calibration, likely indicating overconfidence or underconfidence in their predictions. Examining confidence values, CAI4CAI and PredictED consistently show the lowest confidence scores across all three groups. This aligns with their calibration performance, where CAI4CAI also exhibits lower values, suggesting that these models either lack confidence in their predictions or fail to align well with the ground truth segmentations provided by multiple experts. 

\begin{table}[t]
\fontsize{11}{11}\selectfont
\centering
\begin{tabular}{p{2cm}|c|c|c|c|c|c|c}
  \multirow{2}{2cm}{\textbf{Algorithm}} & \multicolumn{3}{c|}{\textbf{DSC (\%)}} & \multicolumn{3}{c}{\textbf{Confidence (\%)}} \\
  %\usepackage{lineno}
  %\hline
  & \textbf{A} & \textbf{B} &\textbf{C} & \textbf{A} & \textbf{B} & \textbf{C} \\
  \midrule
  \textbf{MedIG} & 96.51 & 97.00 & 90.55 & 98.85 & 98.97 & 95.97 \\
  \textbf{PrAEcision} & 94.67 & 95.40 & 90.06 & 97.77 & 98.03 & 95.85 \\
  \textbf{BreizhSeg} & 95.42 & 96.09 & 86.81 & 98.26 & 97.83 & 95.59 \\
  \textbf{DLAI} & 94.88 & 96.15 & 87.57 & 97.17 & 97.89 & 93.82 \\
  \textbf{BCNAIM} & 93.37 & 93.36 & 85.34 & 96.97 & 97.25 & 93.62 \\
  \textbf{CAI4CAI} & 87.20 & 89.13 & 79.07 & 93.20 & 93.95 & 89.37 \\
  \textbf{PredictED} & 85.69 & 86.78 & 84.93 & 92.67 & 92.84 & 91.73 \\
\end{tabular}
\caption{Group-wise DSC and Confidence for all segmentation algorithms.}\label{tab:groupwise_metrics1}
\end{table}

\begin{table}[t]
\fontsize{11}{11}\selectfont
\centering
\begin{tabular}{p{2cm}|c|c|c|c|c|c}
  \multirow{2}{2cm}{\textbf{Algorithm}} & \multicolumn{3}{c|}{\textbf{ECE ($\times 10^{-3}$)}} & \multicolumn{3}{c}{\textbf{CRPS ($cm^{3}$)}} \\
  %\usepackage{lineno}
  %\hline
  & \textbf{A} & \textbf{B} & \textbf{C} & \textbf{A} & \textbf{B} & \textbf{C} \\
  \midrule
  \textbf{MedIG} & 1.88 & 1.72 & 1.86 & 7.52 & 8.63 & 8.12 \\
  \textbf{PrAEcision} & 2.00 & 2.16 & 2.49 & 10.00 & 12.58 & 8.78 \\
  \textbf{BreizhSeg} & 1.19 & 1.15 & 2.42 & 14.56 & 13.56 & 9.20 \\
  \textbf{DLAI} & 4.82 & 2.59 & 4.37 & 13.56 & 14.91 & 9.62 \\
  \textbf{BCNAIM} & 5.85 & 6.86 & 5.89 & 9.296 & 10.73 & 9.14 \\
  \textbf{CAI4CAI} & 6.26 & 4.04 & 3.36 & 14.68 & 13.89 & 10.20 \\
  \textbf{PredictED} & 5.97 & 7.09 & 6.78 & 28.73 & 28.25 & 21.18 \\
\end{tabular}
\caption{Group-wise ECE and CRPS for all segmentation algorithms.}\label{tab:groupwise_metrics2}
\end{table}

Overall, MedIG demonstrates the most stable performance, exhibiting only minor variations in metrics across groups. In contrast, BreizhSeg and DLAI show greater variability, performing well in Groups A and B but experiencing a notable drop in Group C, suggesting that their robustness is dataset-dependent. PredictED displays the most unstable performance, with significant CRPS fluctuations, indicating that it struggles the most with volume predictions. This is evident from its high variability across nearly all four metrics, as shown in Figure \ref{fig:metrics}.

As expected, Group C proves to be the most challenging test subset, leading to performance declines across all models. An interesting observation is the strong correlation between ECE and CRPS, suggesting that better-calibrated models tend to produce more accurate volume estimations. This relationship is expected, given that predicted volumes are derived by summing probabilities from the probabilistic segmentation, which also contributes to the ECE calculation. A particularly intriguing finding is that, for both CRPS and DSC, performance was higher on group B than on Group A, contradicting the initial assumption that anatomical differences between these two groups would impact segmentation difficulty. This suggests that these anatomical variations may be less influential than previously thought in the segmentation of these organs and, only for Group C, these pathologies might have an effect in the final results.

\subsection{Qualitative results}
\label{subsec4.4}

\begin{figure}
    \centering
    \includegraphics[width=\linewidth]{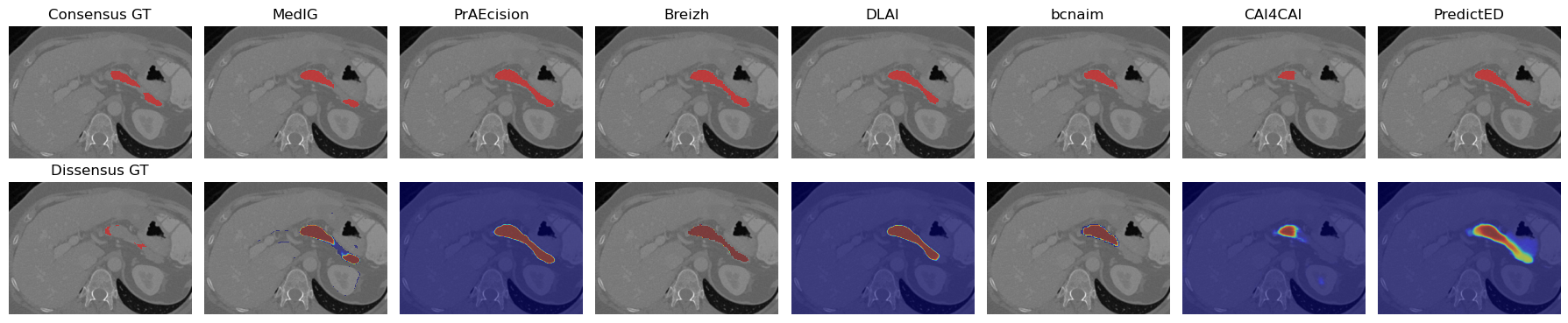}
    \includegraphics[width=\linewidth]{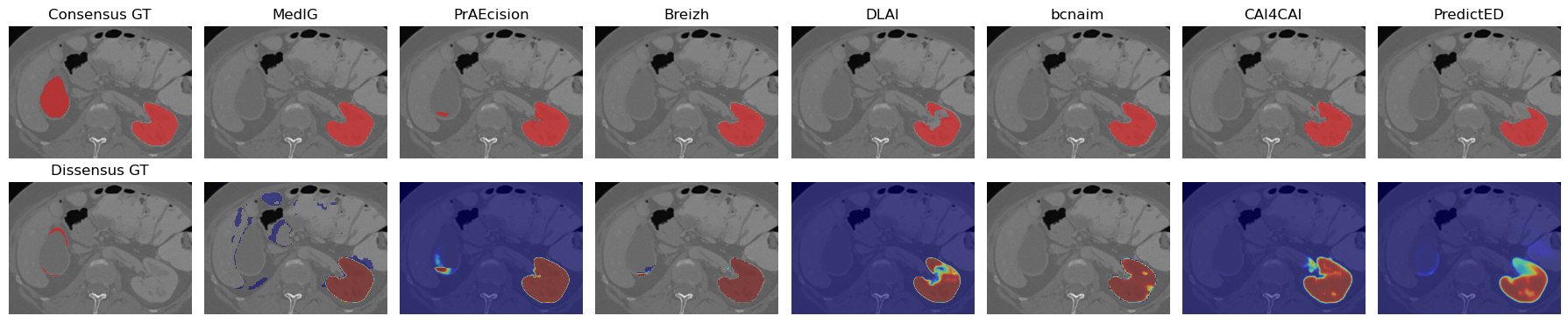}
    \includegraphics[width=\linewidth]{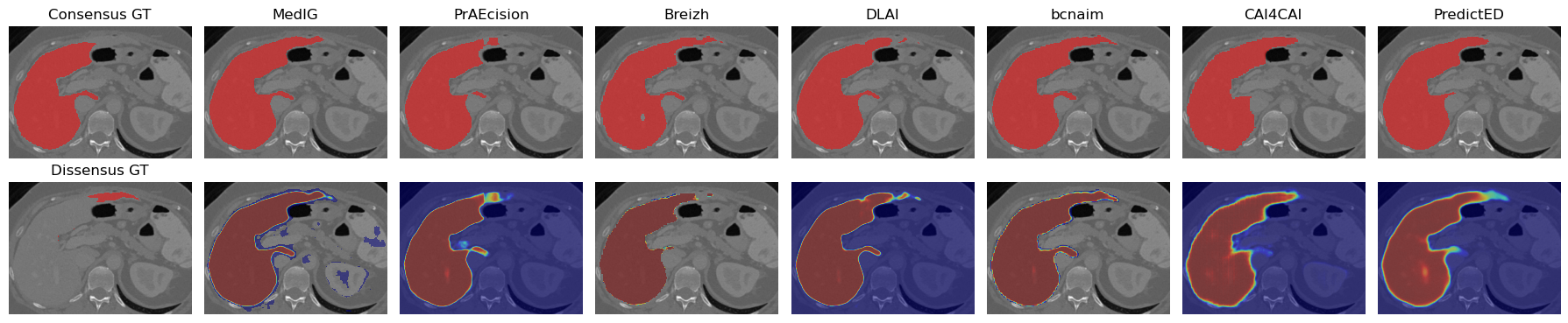}
\caption{Qualitative example of the segmentation produced by the different algorithms (ordered left to right by overall ranking) fare shown for the three target structures: pancreas (rows 1–2), kidneys (rows 3–4), and liver (rows 5–6). For each structure, the top row displays the consensus ground truth (rows 1, 3, 5) and the corresponding binarized segmentations, while the bottom row presents the dissensus ground truth (rows 2, 4, 6) alongside the probabilistic predictions (uncertainty maps).}
\label{fig:qualitative_example}
\end{figure}

We focus our qualitative analysis on a specific test case belonging to Group C (see Figure \ref{fig:qualitative_example}), where the kidneys showed a relatively anomalous shape, as this allows us to gain some insight on the strengths and weaknesses of different models.

Figure \ref{fig:qualitative_example} presents a comparative analysis of organ segmentation results from multiple algorithms of an image belonging to Group C, alongside with its consensus and dissensus GT. Overall, the segmentation quality is high across all methods, with strong agreement between models and the consensus GT. Most algorithms effectively capture the general shape and boundaries of the segmented organs, demonstrating their robustness in standard cases.

The dissensus GT highlights regions where annotators disagreed, often reflected in areas of higher uncertainty. These regions indicate anatomically challenging structures where segmentation ambiguity is more pronounced, likely due to factors such as low contrast, partial volume effects, or anatomical variability. The uncertainty maps further emphasize these discrepancies, particularly in complex structures like the pancreas, where inter-annotator variability is inherently higher.

Interestingly, all algorithms fail to segment one of the kidneys, likely due to its anomalous appearance. 
This failure shows that the models have been predominantly illustrates how models trained on scarce data without large anatomical variability (Group A) tend fo fail to generalize to more complex data (Groups B and C). 
Notably, the only algorithm that managed to segment part of this kidney with relatively high confidence is the PrAEcision, which is the one trained with more additional public datasets, highlighting the importance of diverse training data in handling atypical cases. Additionally, the BreizhSeg algorithm, which used a pre-trained model, also showed a degree of confidence in detecting this kidney anomaly, as reflected in the probabilistic prediction. 
However, PredictED, used a pre-trained ImageNet encoder, which may explain a slight confidence in this anomaly although not reaching the confidence level of PrAEcision or BreizhSeg. 
Finally, BCNAIM, which was trained on the public AbdomenAtlas-8k dataset, failed to detect the anomalous kidney. 
This is likely because areas with low confidence, similar to what was observed with PredictED, appear to have been set to zero due to a heavy post-processing strategy, potentially eliminating even subtle indications of uncertainty. 
These observations indicate that exposure to broader datasets or pre-trained knowledge may improve segmentation performance in cases deviating from standard anatomical presentations.

Among all models, MedIG and PrAEcision exhibit the closest alignment with the consensus GT, showing minimal deviations. Conversely, certain algorithms display greater inconsistencies, particularly in the dissensus GT regions, where segmentation uncertainty is more apparent. Notably, four algorithms (PrAEcision, DLAI, CAI4CAI, and PredictED) consistently predict nonzero values across the entire segmentation map. This behavior directly impacts volume estimation and calibration assessment, potentially leading to systematic biases in predicted organ sizes.

The heatmaps within the dissensus GT further illustrate the extent of disagreement between models and human annotators. These discrepancies are especially pronounced in smaller, less well-defined structures, underscoring the need for improved uncertainty quantification methods. While most models perform well on larger, well-contrasted organs such as the liver, performance variability increases in more challenging anatomical regions, such as the pancreas.

This analysis underscores the importance of multi-annotator ground truth and uncertainty-aware evaluation in medical image segmentation, given the inherent subjectivity in manual annotations.
Instead, robust segmentation models must account for annotation variability and provide reliable confidence estimates, ensuring clinically meaningful predictions. Additionally, our findings suggest that incorporating diverse datasets and leveraging pre-trained models can enhance robustness, particularly when encountering anomalous cases that deviate from the expected anatomical norm.

\subsection{Statistical Analysis}
\label{subsec4.5}

%\textcolor{red}{
The statistical analysis of the results comprises two complementary components: first, an assessment of the statistical significance of performance differences between algorithms; and second, an evaluation of the stability and robustness of the resulting rankings.%}

\begin{figure}
    \centering    \includegraphics[width=\linewidth]{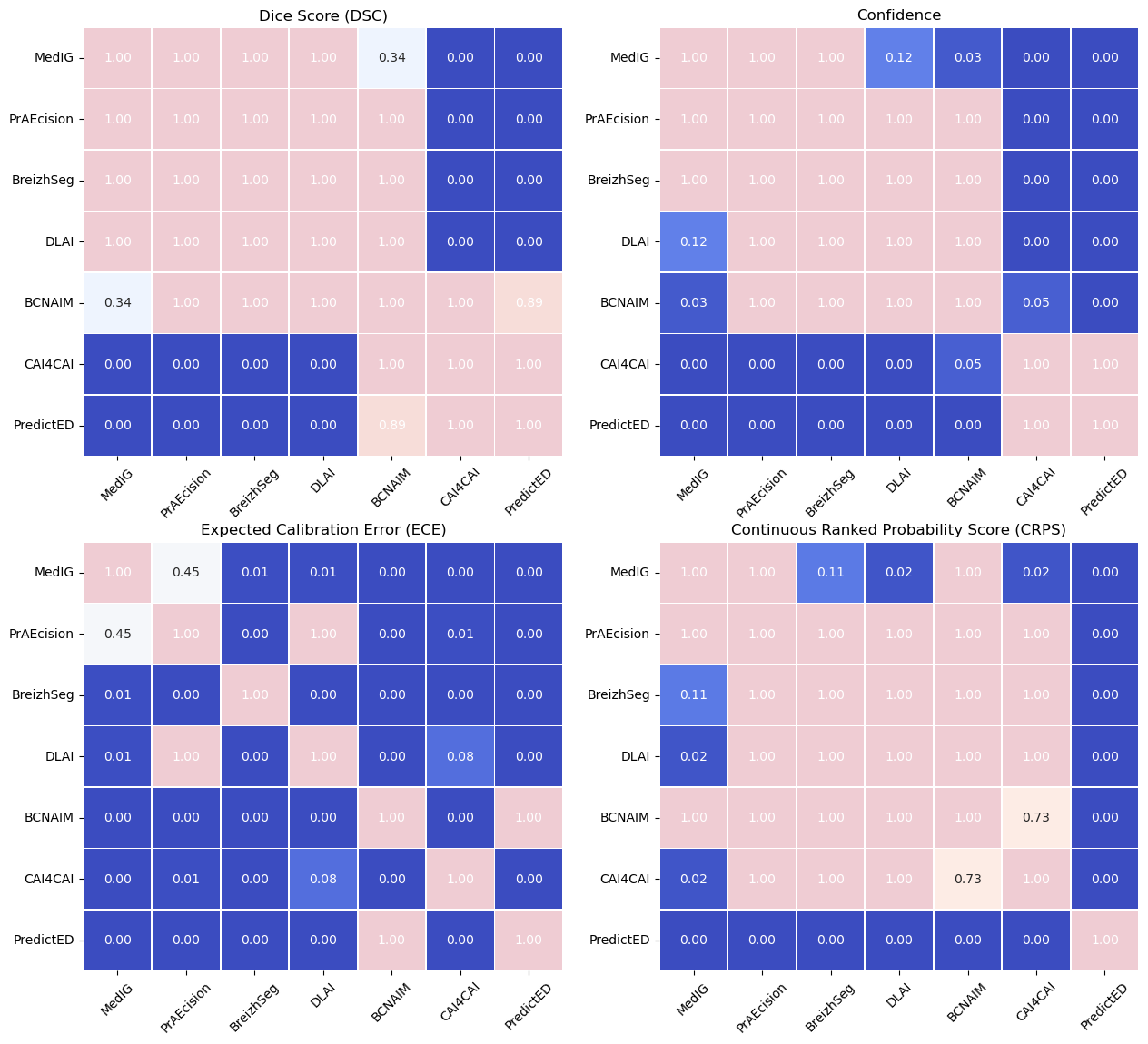}
\caption{Pairwise statistical comparisons (corrected p-values) across segmentation algorithms for four performance metrics: Dice Score (DSC), Confidence, Expected Calibration Error (ECE), and Continuous Ranked Probability Score (CRPS).
Each heatmap shows the results of pairwise Wilcoxon signed-rank tests between methods, with p-values corrected for multiple comparisons using the Holm–Bonferroni method.
Darker cells (\textit{blue}) indicate statistically significant differences ($p<0.05$), while lighter cells indicate non-significant results after correction. Diagonal values are 1.0 by definition (self-comparisons).}
\label{fig:wilcoxon}
\end{figure}

%\textcolor{red}{
Figure \ref{fig:wilcoxon} presents the pairwise statistical significance results based on two-sided Wilcoxon signed-rank tests conducted across all participating methods. The comparisons were performed separately for each evaluation metric, Dice Score (DSC), Expected Calibration Error (ECE), Continuous Ranked Probability Score (CRPS), and confidence reliability, with p-values adjusted using Bonferroni correction to account for multiple testing. In the heatmaps, darker shades correspond to smaller adjusted p-values, indicating statistically significant differences between method pairs.%}

%\textcolor{red}{
This analysis serves to reinforce and contextualize performance-based rankings by identifying which observed differences are statistically robust. 
Notably, the top-performing methods, MedIG, PrAEcision, and BreizhSeg, consistently exhibit statistically significant improvements in calibration and uncertainty-related metrics (ECE, CRPS, and confidence reliability) when compared to lower-ranking teams such as PredictED, CAI4CAI, and BCNAIM. In contrast, the Dice Score matrix reveals fewer statistically significant differences, underscoring that segmentation accuracy alone does not fully capture model quality, highlighting the importance of comprehensive evaluation in multi-rater, uncertainty-aware settings.%}

To analyze the statistical dispersion and robustness of algorithm performance, we employed a bootstrapping procedure to estimate the confidence in metric means and assess the stability of algorithm rankings. Specifically, we performed 500 bootstrap iterations by resampling the test cases with replacement. For each iteration, the mean metric score per algorithm was computed, followed by ranking all algorithms accordingly. From these bootstrap-derived rankings, we calculated the mean rank, standard deviation (as a measure of rank stability), and median rank (to provide a robust central tendency measure) for each algorithm. Additionally, we computed 95\% confidence intervals for the mean ranks using $\pm1.96$ times the standard deviation, offering a statistical indication of ranking reliability. Algorithms were then sorted by their mean rank, with lower values indicating better performance.

\begin{figure}
    \centering
    \includegraphics[width=\linewidth]{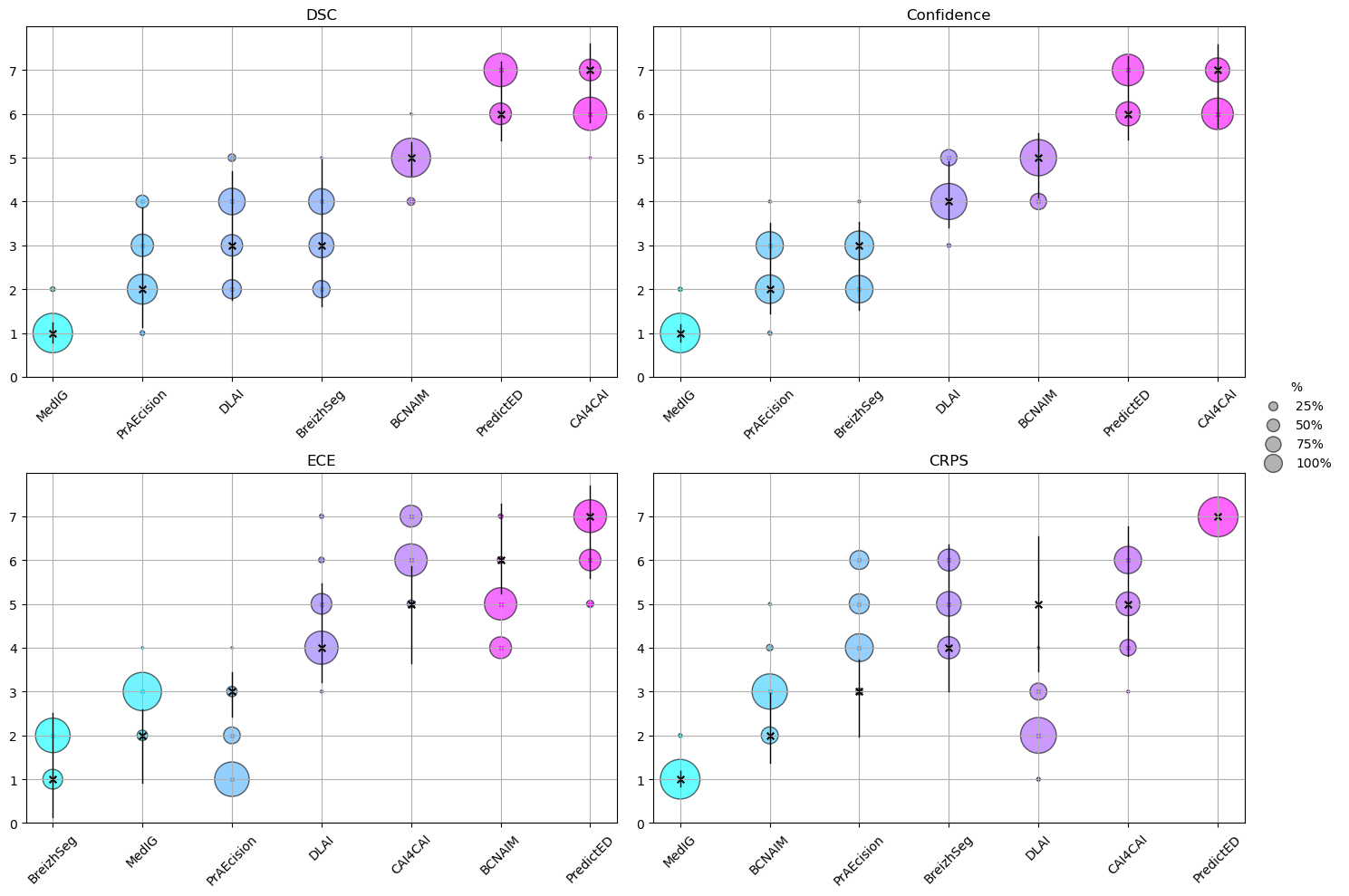}
\caption{Ranking stability for each metric of each algorithm. The size of the bubble refers to the amount of times percentage-wise the algorithm ended up in this position of the ranking after carrying out a bootstrapping process. Black crosses indicate the median rank for each algorithm and black lines indicate the 95\% bootstrap intervals across samples.}
\label{fig:blob_plot}
\end{figure}

Figure \ref{fig:blob_plot} presents the ranking stability of different algorithms across four evaluation metrics: DSC, Confidence, ECE, and CRPS. The ranking distribution for each algorithm is visualized using bubble plots, where the size of each bubble represents the percentage of bootstrap samples in which the algorithm achieved a given rank. Black crosses indicate the median rank, and vertical black lines represent the 95\% bootstrap confidence intervals, highlighting the variability in rankings across different resampled datasets. The x-axis ordering of algorithms varies between subplots because rankings are computed independently for each metric. Unlike a fixed ordering, the x-axis positions are determined by the distribution of rankings and the median rank within each metric. Since algorithms do not perform uniformly across all metrics, their relative positions shift accordingly. For instance, an algorithm that ranks highly in DSC may perform poorly in CRPS, leading to a different placement on the x-axis. This variability highlights the importance of multi-metric evaluations when assessing algorithm performance, as rankings can change significantly depending on the metric considered.

The rankings exhibit varying degrees of stability across the four metrics. For DSC and Confidence, the top-performing algorithm, MediG, consistently achieves rank 1 in most bootstrap samples, as indicated by the large bubble at this position. In contrast, algorithms such as CAI4CAI and PredictED frequently occupy the lower ranks, with relatively small variations in ranking distributions, suggesting consistently weaker performance in these metrics. The spread of rankings, as observed in the confidence intervals, varies across algorithms, indicating that some methods are more stable in their ranking placement than others.

In ECE, the ranking distribution highlights that BreizhSeg and MedIG achieve top positions with high frequency, suggesting strong calibration capabilities. However, an interesting observation is that, for some algorithms, the median ranking (black cross) does not align with the largest bubble, indicating that although a rank is frequently occupied, variability across bootstrap samples influences the final median position, potentially signaling skewed distributions in ranking assignments. For instance, DLAI and BCNAIM show a noticeable shift between the most frequent rank and the median, suggesting occasional fluctuations in ranking.

For CRPS, MedIG and PrAEcision again demonstrate superior performance, consistently ranking among the top algorithms. However, a similar pattern to ECE emerges, where some algorithms have their median ranking positioned differently from the most frequently occupied rank. This effect is particularly evident for BreizhSeg and BCNAIM, where a broader ranking distribution suggests greater sensitivity to bootstrap sampling. Conversely, PredictED ranks the lowest with a large bubble at rank 7, indicating frequent poor performance. The broad confidence intervals for some algorithms suggest that performance variations across different bootstrap samples are non-negligible.

%\textcolor{red}{
It is worth noting that the plotted bubbles represent the mean aggregated rank for each method, whereas the 95\% confidence intervals are obtained by bootstrapping per-case scores before rank aggregation. Because the rank transformation is nonlinear, the mean of the bootstrapped ranks does not necessarily match the mean aggregated rank. This can lead to apparent misalignments between the bubble centers and their corresponding confidence intervals, which reflect uncertainty in the underlying per-case scores rather than in the aggregated ranks.
%}

The ranking stability analysis offers valuable insights into the robustness and consistency of different algorithms across multiple evaluation metrics. MedIG emerges as the most stable and consistently high-performing algorithm, whereas CAI4CAI and PredictED exhibit lower and more variable rankings. The variability in bootstrap confidence intervals highlights which algorithms maintain stable performance and which are more sensitive to dataset variations. Notably, the discrepancies between the most frequent rank and the median in ECE and CRPS suggest skewed ranking distributions, emphasizing the need to consider both ranking stability and variability. Furthermore, the differences in x-axis ordering across metrics underscore the diverse behavior of algorithms, reinforcing the importance of comprehensive multi-metric evaluations for assessing algorithm performance. %These findings are essential 

\section{Discussion}
\label{sec5}

This paper presents the results of the CURVAS Challenge at MICCAI 2024. All submissions were deep learning-based, specifically UNet architectures, reaffirming DL’s dominance in abdominal medical image segmentation. All methods achieved comparable average performance, although a more nuanced analysis revealed hidden trends and different scenarios where some methods performed better than others.

\subsection{Best Performing methods}
\label{subsec5.1}

The top performing teams are the MedIG, PrAEcision and BreizhSeg. 
The three of them are based in nnUnet models but for the inference and uncertainty quantification, MedIG and PrAEcision do not explicitly focus on uncertainty quantification but rather focus on accuracy and segmentation quality. 
MedIG uses the 3D full-resolution version incorporating post-processing to meet storage restrictions and speed up processing. 
PrAEcision uses the 3D low-resolution nnUNet to meet the challenge’s inference time constraints since the low-resolution model sacrifices some accuracy but ensures faster processing. 
BreizhSeg uses a pre-trained Total Segmentator (TS) with an Adaptable Bayesian Neural Network (ABNN) to provide uncertainty quantification, leveraging Monte Carlo sampling to generate confidence maps and improve robustness in the face of limited data. 
This is a more computationally intensive approach, requiring multiple predictions to estimate uncertainty, which may slow inference but provides valuable uncertainty information.

The three top-ranked strategies also engage in preprocessing steps to standardize the data, such as resampling and standardizing voxel size (BreizhSeg) or optimizing input sizes and formats for faster inference (MedIG and PrAEcision) but the approaches to handle the variabilities in the annotations differ. 
MedIG uses STAPLE to combine multiple annotations into a single consensus segmentation by iterating on sensitivity and specificity estimates; PrAEcision incorporates a wider variety of training datasets, which indirectly handles the variability by exposing the model to more diverse data from different sources; and BreizhSeg relies on a Bayesian approach (ABNN) and uses TS to handle uncertainty quantification, focusing on the model’s ability to estimate uncertainty, which is an advanced approach to handle limited training data and generalization, while also helping to deal with annotation variability.

In summary, all three teams show common preference for nnU-Net's proven effectiveness in medical imaging segmentation, but extending it with different mechanisms to make it more robust to data scarcity and aware of annotation uncertainty as well as employing different strategies to handle annotation variability, training data, model selection, and inference efficiency. 
MedIG focuses on post-processing to optimize nnU-Net, PrAEcision emphasizes fast inference with a low-res nnU-Net and expanded data, and BreizhSeg integrates uncertainty quantification through a Bayesian approach to improve segmentation with limited data.

%\textcolor{red}{
We have used pairwise Wilcoxon signed-rank tests across metrics to study whether there are statistically significant differences between the performances of the methods. 
We found that, based on the corrected p-value matrices (see Figure \ref{fig:wilcoxon}), there is no statistically significant difference between the top-performing methods, MedIG, PrAEcision, and BreizhSeg, across any of the four evaluation metrics, suggesting their performance is comparable and robust with respect to both segmentation accuracy and uncertainty calibration. 
However, performance differences between teams translated into statistically significant improvements, particularly in terms of uncertainty calibration. 
We observed that top-performing methods consistently outperformed lower-ranking teams such as PredictED, BCNAIM, DLAI, and CAI4CAI ($p<0.05$ after Bonferroni correction).%}

%\textcolor{red}{
Notably, these significance tests revealed that differences in Dice score were less pronounced, suggesting that segmentation accuracy alone does not capture the full spectrum of model quality. Instead, calibration metrics better reflected performance differences among the top contenders, reinforcing the importance of evaluating not just what the model predicts, but how confident it is in those predictions. 
This confirms that uncertainty-aware models can meaningfully outperform others even when segmentation accuracy appears comparable which shows the necessity of a holistic evaluation strategy that goes beyond pixel-wise overlap to assess trustworthiness and calibration in clinical AI systems. This statistical analysis complements the performance-based rankings by identifying where observed differences are robust and reproducible.%} %, reinforcing that calibration and uncertainty estimates are critical to model quality beyond segmentation accuracy alone.

\subsection{Clinical applications}
\label{subsec5.3}

Understanding and analyzing variability is essential in medical imaging, where discrepancies in expert annotations or algorithmic outputs can significantly impact diagnosis and patient care. Acknowledging this uncertainty and incorporating multiple perspectives enhances both the accuracy and reliability of clinical decisions. Several clinical applications could benefit from the integration of variability studies, particularly in areas such as radiology, oncology, and pathology. %\textcolor{red}{
For instance, in oncology, understanding variability in imaging data could improve the assessment of treatment response, as different radiologists or automated systems may interpret changes in tumor morphology differently. In radiology, variability studies could help refine tumor segmentation techniques, allowing for more accurate measurements of tumor size and location, which are critical for treatment planning and monitoring.
For example, when radiologists assess vascular invasion in the pancreas—a critical factor for determining tumor resectability—the evaluation heavily relies on clearly defining the tumor’s boundaries. However, these boundaries often vary between experts due to image ambiguity and low contrast. A calibrated model could flag regions with high uncertainty, such as the tumor margins near vessels, and explicitly alert the clinician. This would allow radiologists to consider not only the model's prediction but also the reliability of that prediction, enabling more nuanced interpretations.
Ultimately, integrating such uncertainty-aware visual cues into diagnostic workflows could improve both trust in AI systems and decision quality in borderline or ambiguous cases.%}
By quantifying these variations, clinicians could better account for uncertainties when making decisions about patient management. In pathology, variability studies could be applied to the classification of tissue samples, ensuring that multiple experts or algorithms arrive at consistent results, reducing the risk of misclassification in diagnosing diseases such as cancer. 
Furthermore, integrating variability analysis into clinical decision support systems could enhance personalized medicine, helping tailor treatment strategies based on the degree of uncertainty in individual cases. In the end, applying these insights to clinical practice would lead to more robust, transparent, and reliable diagnostic processes, providing clinicians with the information needed to make better-informed, patient-centered decisions.

Despite the increasing awareness of this issue, there remains a gap in applying these insights to real-world clinical settings. There is a clear need to develop practical methods for leveraging variability information in ways that enhance the clinical workflow and support clinicians in making more informed decisions. Additionally, it is essential that the concept of non-golden standard ground truths becomes more widely accepted within both the technical and medical communities. 

\subsection{Lessons learned and limitations}
\label{subsec5.4}

An important consideration that emerged from this challenge is the choice of evaluation metrics used to assess algorithm performance. These metrics should be optimized not only for clinical relevance but also for computational efficiency, particularly when handling large-scale volumetric images. Furthermore, incorporating ranking variability into the evaluation framework would enable a more nuanced and robust interpretation of algorithm performance, ultimately leading to more reliable final rankings.

One of the main limitations of this challenge lies in the dataset size. While we prioritized acquiring multiple expert annotations per CT scan—a critical factor for assessing inter-rater uncertainty—this came at the cost of a reduced number of total scans. Future editions of the challenge would benefit from engaging multiple clinical centers to increase the diversity and variability of the dataset. Additionally, more rigorous quality assurance procedures should be implemented during the data preparation phase. In the initial version of the dataset published on Zenodo, a few instances of corrupted metadata were identified. Although these issues were promptly detected and corrected, we acknowledge the need for stricter validation to avoid similar occurrences in future releases.

\section{Conclusion and future directions}
\label{sec6}

This study highlights the importance of evaluating medical image segmentation models beyond traditional accuracy metrics by incorporating multi-annotator ground truth, uncertainty estimation, and ranking stability. Our findings underscore key challenges in medical image segmentation and provide insights into how models can be improved for real-world applications.  

First, there is no single "gold standard" in medical image segmentation due to the inherent subjectivity in manual annotations. Multi-annotator ground truth is essential for capturing the variability in expert opinions and ensuring a more comprehensive evaluation of segmentation performance. Future models should account for this variability rather than relying solely on single-label annotations.  
Second, uncertainty estimation plays a critical role in medical image segmentation, particularly in anatomically complex or low-contrast regions where inter-annotator disagreement is high. Models that incorporate probabilistic outputs or Bayesian approaches demonstrate improved robustness by providing reliable confidence estimates. This suggests that future segmentation frameworks should integrate uncertainty quantification techniques to better inform clinical decision-making. Furthermore, model calibration is crucial for ensuring that predicted confidence scores accurately reflect the true likelihood of correct segmentation. Poorly calibrated models may produce overconfident or underconfident predictions, leading to misinterpretation of segmentation outputs. Our results indicate that models with superior calibration not only provide more reliable confidence estimates but also tend to yield more accurate volume estimations, which is essential for clinical applications requiring precise organ measurements.
Third, the choice of training data and pre-trained models significantly impacts segmentation performance and generalizability. Algorithms trained on diverse datasets, particularly those incorporating public data and pre-trained encoders, tend to handle atypical cases more effectively. This highlights the importance of dataset diversity and transfer learning in medical imaging applications.  
Finally, multi-metric evaluation is necessary to fully understand model performance. While segmentation accuracy (e.g., DSC) remains a primary benchmark, other metrics such as calibration, ranking stability, and volume estimation accuracy provide deeper insights into model reliability. Evaluating models through a comprehensive set of criteria ensures that they perform robustly across different anatomical structures and patient populations.  

A major challenge for future CURVAS editions will be not only identifying new clinical applications for concepts such as variability, uncertainty, and calibration, but also determining how to effectively communicate these aspects to clinicians. 
Conveying the importance of these elements is essential to ensure diagnostic decision-making in a more comprehensive understanding of both data and model limitations. This involves moving beyond presenting only point predictions to also providing insights into prediction confidence and expert disagreement. 
In parallel, the development of segmentation models should prioritize generalizability and the generation of well-calibrated, uncertainty-aware predictions. 
%\textcolor{red}{
While our dataset provides a valuable benchmark, its size and source may limit generalizability across institutions or scanner types. 
Future work should explore strategies to mitigate domain shift, such as incorporating domain adaptation techniques or fine-tuning models on site-specific data. %} %These approaches could help ensure robust performance and well-calibrated predictions in diverse clinical environments.
To achieve this, future efforts should emphasize the integration of diverse datasets as well, use of pre-trained models, and adoption of advanced calibration techniques—alongside the inclusion of expert discrepancies in both training and evaluation pipelines—to build more trustworthy and clinically relevant AI systems.

\section*{CRediT authorship contribution statement}
\label{sec7}

\textbf{Meritxell Riera-Marín:} Conceptualization, Methodology, Software, Investigation, Writing - Original Draft, Writing - Review \& Editing, Data Curation, Formal analysis, Validation.
\textbf{Sikha O K:} Conceptualization, Software, Visualization. 
\textbf{Júlia Rodríguez-Comas:} Conceptualization, Funding acquisition. 
\textbf{Matthias Stefan May:} Conceptualization, Data Curation, Funding acquisition. 
\textbf{Joy-Marie Kleiss:} Data Curation, Investigation, Resources. 
\textbf{Anton Aubanell:} Data Curation. 
\textbf{Andreu Antolin:} Data Curation. 
\textbf{Javier García-Lópeza:} Supervision, Writing - Review \& Editing, Funding acquisition. 
\textbf{Miguel A. González-Ballester:} Supervision, Writing - Review \& Editing, Funding acquisition. 
\textbf{Adrián Galdrán:} Conceptualization, Supervision, Writing - Review \& Editing, Funding acquisition. 

\textbf{Zhaohong Pan:} Software, Investigation, Writing - Review \& Editing.
\textbf{Xiang Zhou:} Software, Investigation.
\textbf{Xiaokun Liang:} Software, Investigation.
\textbf{Franciskus Xaverius Erick:} Software, Investigation, Writing - Review \& Editing.
\textbf{Andrea Prenner:} Software, Investigation, Writing - Review \& Editing.
\textbf{Cédric Hémon:} Software, Investigation, Writing - Review \& Editing.
\textbf{Valentin Boussot:} Software, Investigation, Writing - Review \& Editing.
\textbf{Jean-Louis Dillensegers:} Software, Investigation.
\textbf{Jean-Claude Nunes:} Software, Investigation.
\textbf{Abdul Qayyum:} Software, Investigation, Writing - Review \& Editing.
\textbf{Moona Mazher:} Software, Investigation.
\textbf{Steven A Niederer:} Software, Investigation.
\textbf{Kaisar Kushibar:} Software, Investigation, Writing - Review \& Editing.
\textbf{Carlos Martín-Isla:} Software, Investigation.
\textbf{Petia Radeva:} Software, Investigation.
\textbf{Karim Lekadir:} Software, Investigation.
\textbf{Theodor Barfoot:} Software, Investigation, Writing - Review \& Editing.
\textbf{Luis C. Garcia Peraza Herrera:} Software, Investigation.
\textbf{Ben Glocker:} Software, Investigation.
\textbf{Tom Vercauteren:} Software, Investigation.
\textbf{Lucas Gago:} Software, Investigation, Writing - Review \& Editing.
\textbf{Justin Englemann:} Software, Investigation, Writing - Review \& Editing.

\section*{Code availability}
\label{sec8}

To support users with the evaluation of the algorithms, we have made the evaluation code used in this manuscript publicly available in the Challenge’s GitHub repository \cite{curvas2024_github}.

\section*{Declaration of Competing Interest}
\label{sec9}

The authors confirm that they have no known financial or personal conflicts of interest that could have influenced the work presented in this paper.

\section*{Data availability}
\label{sec10}

All data has been uploaded on Zenodo \cite{CURVAS_Zenodo}.

\section*{Acknowledgments}
\label{sec11}

The authors gratefully acknowledge funding from the following sources:

This work was supported by the Catalan Government through the "Doctorats Industrials" program, specifically the industrial doctorate AGAUR 2021-063, in collaboration with Sycai Technologies SL. 
The challenge is part of the Proyectos de Colaboración Público-Privada (CPP2021-008364), funded by MCIN/AEI/10.13039/501100011033 and co-financed by the European Union through the NextGenerationEU/PRTR initiative. 
Addiotional funding was provided by the "NUM 2.0" project (FKZ: 01KX2121) and the Eureka Eurostars-3 joint program (1661 DARE-KPL), co-funded by the Horizon Europe Research and Innovation Framework Programme of the European Union (FKZ: 347 01QE2249C).

A.G. is supported by grant RYC2022-037144-I, funded by \newline MCIN/AEI/10.13039/501100011033 and co-financed by FSE+.

The work of Z.P., X.Z. and X.L. is supported by grants from the National Key Research and Develop Program of China (2023YFC2411502) and the National Natural Science Foundation of China (82202954).

A.P. and F.E. received support from the High-tech agenda Bavaria and HPC resources provided by the Erlangen National High Performance Computing Center (NHR@FAU) of the Friedrich-Alexander-Universität Erlangen-Nürnberg (FAU) under the NHR project b180dc. NHR@FAU hardware is partially funded by the German Research Foundation (DFG) - 440719683. Support was also received from the ERC project MIA-NORMAL 101083647 and DFG KA 5801/2-1, INST 90/1351-1, and 512819079.

K.K. holds the Juan de la Cierva fellowship with reference number FJC2021-047659-I.

V.B. was supported by the Brittany Region through its Allocations de Recherche Doctorale framework and by the French National Research Agency as part of the VATSop project (ANR-20-CE19-0015). C.H. was supported through a PhD scholarship grant from Elekta AB.

\section*{Supplementary material}
%\label{sec12}

The detailed challenge proposal is available on Zenodo \cite{CURVAS_proposal}.

%\appendix
%\section{Example Appendix Section}
%\label{app1}

%\textbf{Put the documentation provided by all the participants.}


%%\bibliographystyle{elsarticle-num} 
%%\bibliography{sample.bib}
\bibliographystyle{plain}
\bibliography{main}

\end{document}